%% file: main.tex
\documentclass[conference]{IEEEtran}
\IEEEoverridecommandlockouts
\usepackage{cite}
\usepackage{amsmath,amssymb,amsfonts}
\usepackage{algorithmic}
\usepackage{graphicx}
\usepackage{textcomp}
\usepackage{xcolor}
\usepackage[colorlinks=true, urlcolor=blue, linkcolor=red]{hyperref}
\usepackage{algorithm}
\usepackage{url}
\usepackage{bbm}
\usepackage{bm}
\usepackage{subcaption}
\usepackage[export]{adjustbox}
\usepackage{mathtools}
\usepackage{multirow}

\def\BibTeX{{\rm B\kern-.05em{\sc i\kern-.025em b}\kern-.08em
    T\kern-.1667em\lower.7ex\hbox{E}\kern-.125emX}}
\begin{document}

\title{\input{Texts/title.tex}
}
\author{\IEEEauthorblockN{Zheng Liu}
\IEEEauthorblockA{
\textit{University of Illinois at Chicago}\\
zliu212@uic.edu}
\and
\IEEEauthorblockN{Xiaohan Li}
\IEEEauthorblockA{
\textit{University of Illinois at Chicago}\\
xli241@uic.edu}
\and
\IEEEauthorblockN{Philip S. Yu}
\IEEEauthorblockA{
\textit{University of Illinois at Chicago}\\
psyu@uic.edu}

}

\maketitle

\begin{abstract}
\input{Texts/0-Abstract}
\end{abstract}


\input{Texts/1-Intro}

\input{Texts/2.1-Prelim}

\input{Texts/2.2-Methods}
\input{Texts/3-Exp}
\input{Texts/4-RelatedWorks}

\input{Texts/5-Conclusion}

\bibliographystyle{IEEEtran}

\bibliography{main}

\end{document}

%% file: Texts/title.tex
\title{A Counterfactual Fair Model for Longitudinal Electronic Health Records  via Deconfounder}

%% file: Texts/0-Abstract.tex
The fairness issue of clinical data modeling, especially on Electronic Health Records (EHRs), is of utmost importance due to EHR's complex latent structure and potential selection bias. It is frequently necessary to mitigate health disparity while keeping the model's overall accuracy in practice. However, traditional methods often encounter the trade-off between accuracy and fairness, as they fail to capture the underlying factors beyond observed data. 
To tackle this challenge, 
we propose a novel model called Fair Longitudinal Medical Deconfounder (FLMD) \footnote{ 
\href{https://anonymous.4open.science/r/ICDM_FLMD-C223}{https://anonymous.4open.science/r/ICDM\_FLMD-C223}
} 
that aims to achieve both fairness and accuracy in longitudinal Electronic Health Records (EHR) modeling. Drawing inspiration from the deconfounder theory, FLMD employs a two-stage training process.
In the first stage, FLMD captures unobserved confounders for each encounter, which effectively represents underlying medical factors beyond observed EHR, such as patient genotypes and lifestyle habits. This unobserved confounder is crucial for addressing the accuracy/fairness dilemma.
In the second stage, FLMD combines the learned latent representation with other relevant features to make predictions. By incorporating appropriate fairness criteria, such as counterfactual fairness, FLMD ensures that it maintains high prediction accuracy while simultaneously minimizing health disparities.
We conducted comprehensive experiments on two real-world EHR datasets to demonstrate the effectiveness of FLMD. Apart from the comparison of baseline methods and FLMD variants in terms of fairness and accuracy, we assessed the performance of all models on disturbed/imbalanced and synthetic datasets to showcase the superiority of FLMD across different settings and provide valuable insights into its capabilities.

%% file: Texts/1-Intro.tex
\section{Introduction}
Machine learning applications have greatly improved the medical decision-making processes in accuracy, especially through the modeling of electronic health records (EHRs). However, there is a growing concern regarding the potential for ML models to exacerbate health disparities among patients with sensitive attributes such as gender and ethnicity \cite{pfohl2021empirical,goodman2018machine}. As demonstrated by empirical studies, ML models may result in less accurate or biased predictions \cite{lwowski2021risk,samorani2021overbooked,kim2022countering} for racial minorities.
As a result, the examination of fairness in EHR modeling has become a subject of increasing attention in the ML domain.

\begin{figure}[t]
  
    \begin{subfigure}[]{0.24\textwidth}
        \centering
\includegraphics[height = 1in]{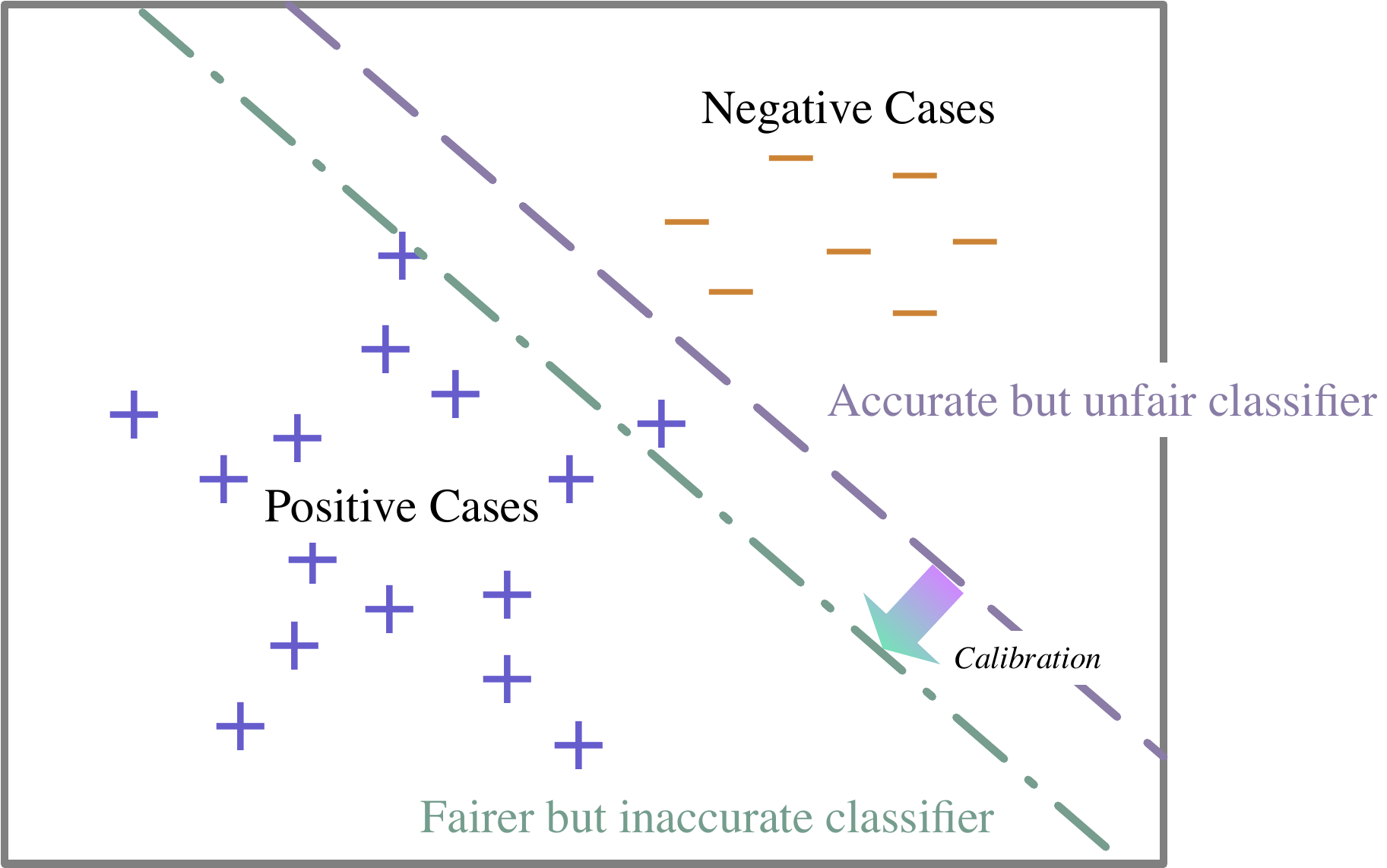}
        \caption{Traditional fairness \\ methods causes accuracy/\\fairness dilemma.}
        \label{fig:toy1}
    \end{subfigure}
    \begin{subfigure}[]{0.24\textwidth}
        \centering
\includegraphics[height = 1in]{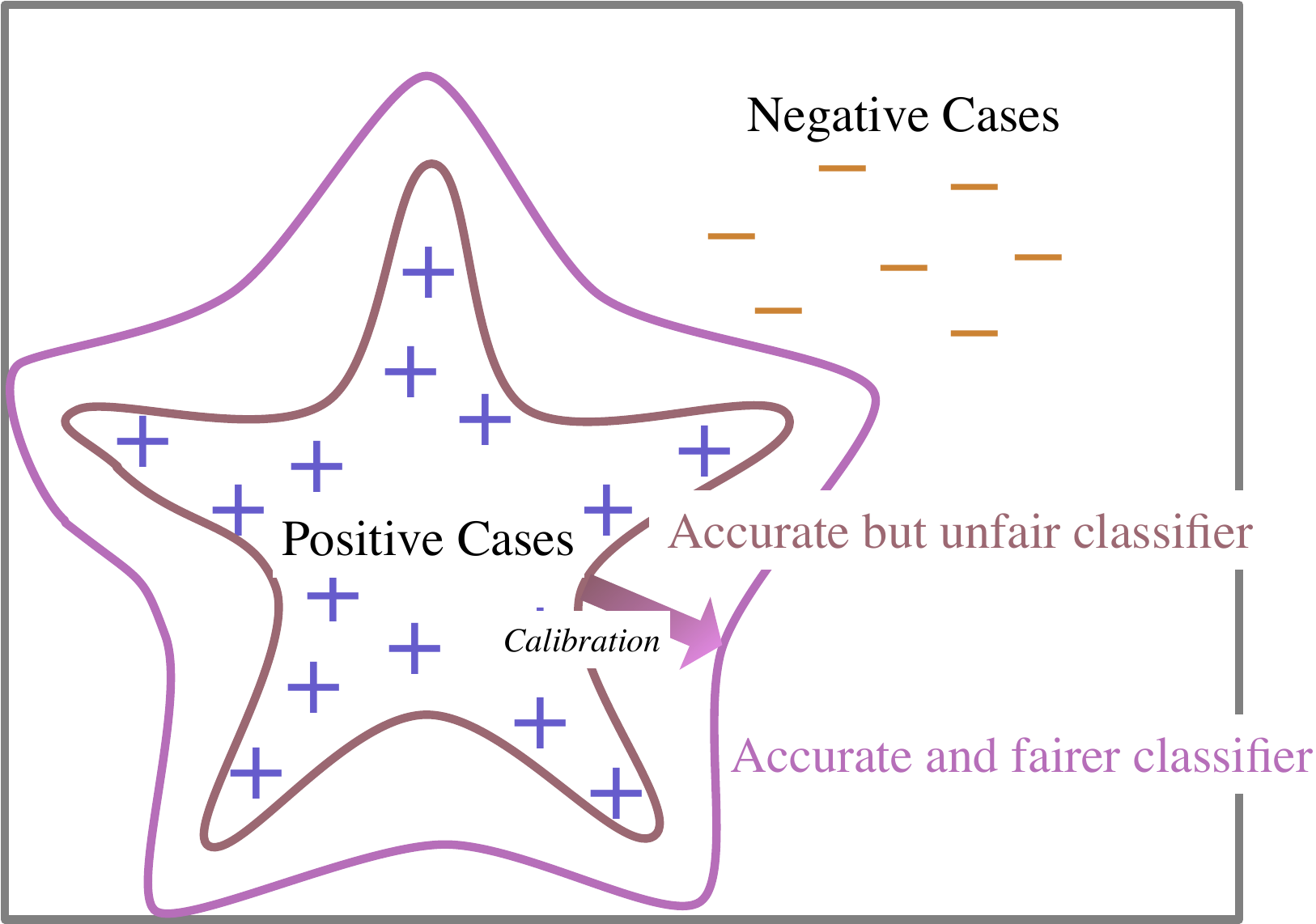}
        \caption{Achieve fairness without accuracy loss by understanding data latent structure.} 
        \label{fig:toy2}
    \end{subfigure}
\caption{An example of how a binary classifier achieves fairness with/without accuracy loss. Here the fairness of a model is defined as the feature space is divided into equal areas. In Figure \ref{fig:toy1}, regardless of how the line (decision boundary) is translated, the classifier fails to divide the feature space into two roughly equal areas while maintaining correct classifications.
In contrast, the Figure \ref{fig:toy2} classifier achieves accuracy and fairness by learning from the distribution of positive samples as a star.}  
\label{fig1}
\end{figure}

Despite the recent progress in ensuring fairness in ML modeling \cite{pleiss2017fairness,louizos2015variational,sattigeri2019fairness}, there are still challenges that hinder the direct application of these methods to EHR modeling. The first challenge comes from the complex latent structure of EHR.
Apart from the clinical observations (features) and outcomes (labels), most EHR datasets contain demographics that are also useful for prediction. From a fairness standpoint, demographics are considered sensitive attributes that require a fair approach when considering their influence. Furthermore, from a causal inference perspective, demographics can act as confounders since they impact both clinical observations and outcomes. They can potentially disrupt the correlation within the data and introduce bias into the model. Therefore, addressing the impact of demographics presents a significant challenge for models aiming to achieve fairness and mitigate bias in EHR modeling. 
Additionally, due to limitations and the cost of health data collection, many other confounders, such as patient genotypes and lifestyle habits, may go unobserved and be excluded from the dataset.
These unobservable confounders can also affect the clinical features we observe and need to be addressed. 
The second challenge comes from potential bias in medical data collection. Due to the high cost and limitation of questionnaires in data collection, observed EHRs are often imbalanced and biased. They may exhibit the missing-not-at-random character, and the sampled distribution may deviate from the true distribution. 
Due to the above issues, most conventional fair methods result in undesirable trade-offs between accuracy and fairness \cite{pfohl2022net,pfohl2021empirical}.

We argue that the accuracy/fairness dilemma encountered by previous fairness approaches stems from their failure to account for the inherent causal structure present in the data. 
An example in Figure \ref{fig1} illustrates our motivation. In Figure \ref{fig1}, a binary classification is performed on a two-dimensional space, where a fair classifier is defined as having a decision boundary that evenly divides the space into two equal areas, and the fairness calibration is defined as the translation of decision boundaries.
The linear classifiers depicted in Figure \ref{fig:toy1} neglect data latent structure. Therefore, even with translation, it is unable for a classifier to achieve fairness and accuracy simultaneously. 
In Figure \ref{fig:toy2}, by recognizing the shape formed by all positive cases, which resemble a star, an accurate but unfair classifier (represented in brown) is initially constructed. Subsequently, the classifier's fairness is improved by expanding the decision boundary without loss of accuracy.
This example highlights the significance of understanding the underlying knowledge about data distribution in fair model development. By doing so, we are more likely to address the accuracy/fairness dilemma and construct a model with enhanced generalization ability.

In this paper, we present a novel model named Fair Longitudinal Medical Deconfounder (FLMD) to learn accurate and highly generalizable knowledge from longitudinal EHR. Based on the above motivation, FLMD consists of two distinct training stages. In the first stage, we adopt a deep generative model to capture the unobserved confounders for each encounter in a patient's EHR. This representation captures the underlying medical factors in EHR, such as patient genotypes and lifestyle habits, and is useful for providing high-quality predictions while keeping fairness. In the second stage, FLMD combines the learned latent embeddings with other relevant features as input for a predictive model. By incorporating
appropriate fairness criteria, such as counterfactual fairness \cite{kusner2017counterfactual},
FLMD ensures that it maintains high prediction accuracy while
simultaneously minimizing health disparities.

We conduct experiments on two real-world datasets to evaluate the performance of FLMD in multiple aspects. First, we compare the performance of FLMD with other baseline models in terms of accuracy and fairness in clinical prediction. Second, we evaluate the models' performance on disturbed/imbalanced data to test the generalization ability of all models. Finally, we evaluate the performance of FLMD on synthetic datasets to see if FLMD (or other baseline latent factor models) can substantially capture the existence of unobserved confounders. And the conclusions of all experiments are listed in the following:

\begin{itemize}
    \item The combination of FLMD and counterfactual fairness outperforms all other baseline methods and FLMD variants in mitigating health disparity while maintaining accurate predictions.
    \item The superior performance of a model on disturbed or imbalanced data, which indicates higher generalization ability, also translates into better fairness predictions and less accuracy sacrifice.
    \item Improved latent model capability in capturing unobserved confounders leads to better performance as well.
\end{itemize}

In summary, in this paper, our contribution is listed as follows:

\begin{itemize}
    \item We introduce a Structural Causal Model (SCM) to represent the underlying latent structure of longitudinal Electronic Health Records (EHR). By framing the fairness issue within the causal inference framework, we are able to address fairness concerns without accuracy loss.
    
    \item We propose a two-stage model called Fair Longitudinal Medical Deconfounder (FLMD). The first stage of FLMD focuses on learning valuable underlying knowledge about the unobserved confounders present in EHR data, while the second stage focuses on effective prediction modeling.

    \item By incorporating the concept of counterfactual fairness, FLMD demonstrates effective mitigation of health disparities while providing highly accurate predictions. Additional experiments highlight the superior performance of FLMD compared to other baseline models under various experiment settings and provide valuable insights for further studies.

\end{itemize}

%% file: Texts/2.1-Prelim.tex
\begin{figure}
        \centering
        \includegraphics[height = 1.1in]{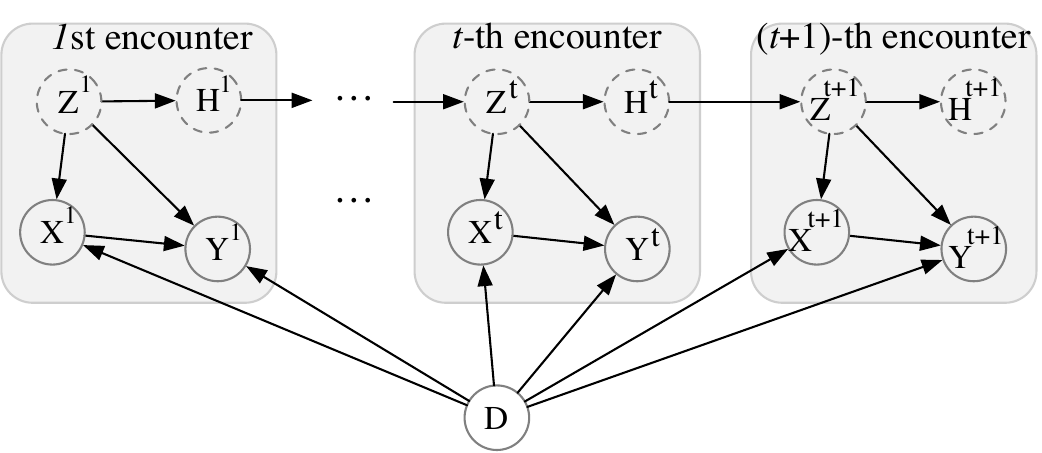}
        \caption{The SCM of a patient's EHRs.} 
        \label{fig:scm}
\end{figure}

\section{Preliminaries}

\subsection{Motivation}

In the field of machine learning, the trade-off between the accuracy and fairness of models is a well-documented phenomenon and has been observed in both theoretical and applied studies \cite{zliobaite2015relation,cooper2021emergent,radovanovic2019making}. It poses a significant challenge to the deployment of AI systems in real-world applications. The trade-off can severely limit the utility of these systems and impede the further advancement of the fairness paradigm.

In this paper, we argue that the accuracy/fairness dilemma in EHR modeling stems from a lack of generalization ability. When different fairness paradigms are applied, they do not interfere with the latent structure of the original data. Therefore, if a model's performance declines significantly after applying fairness interventions, it is likely to be statistically biased and limited in its ability to accommodate various observational distributions. 
Recent studies have connected the generalization ability of models with causal inference theories\cite{bengio2019meta,ke2019learning,xia2021causal}. These studies suggest that models built on accurate causal structures demonstrate superior generalization abilities and increased robustness when faced with distribution shifts. By explicitly incorporating the specific causal structure found in EHRs, we aim to develop a model that enhances generalization ability and potentially overcomes the trade-off between accuracy and fairness.

In order to address the unobserved confounders during this process, a deconfounder \cite{wang2019blessings,wang2020towards} is leveraged in FLMD.
The deconfounder is a two-stage approach designed to derive unbiased estimations in observational studies with multiple causes.
Its first stage involves an unsupervised latent factor model to learn the unobserved confounders. In EHR modeling, this model is employed to capture the underlying medical factors from clinical observations. The results obtained in this stage, if effectively leveraged, can prove to be valuable for subsequent predictions.
Its second stage is a predictive model, which utilizes the outcomes from the first stage to derive final predictions.

\subsection{Problem Formulation}
Here we formulate the problem of longitudinal Electronic Health Record (EHR) modeling. Please note that bold uppercase letters (e.g., $\bm X$) are used to denote random variables in causal models, and lowercase letters (e.g., $\bm x$) are used to denote corresponding data or embedding vectors.

Let $\mathcal{D} = \{\mathcal{E}^{(i)}\}_{i=1}^N$ be the EHRs of $N$ patients.
The EHR of the $(i)$-th patient in the dataset can be denoted as $\mathcal{E}^{(i)}= \{\bm d^{(i)}, \mathcal{X}^{(i)}, \mathcal{Y}^{(i)}\}$. Here, $\bm d^{(i)}$ denotes \textbf{demographics};
$\mathcal{X}^{(i)} = \{\bm x^{(i,1)}, \bm x^{(i,2)}, \cdots, \bm x^{(i,T{(i)})} \}$ and $\mathcal{Y}^{(i)} = \{y^{(i,1)}, y^{(i,2)}, \cdots, y^{(i,T{(i)})} \}$ are two sets containing clinical \textbf{features} and \textbf{labels} of $T{(i)}$ encounters, respectively. We assume the prediction task is binary classification, and thus  the elements in $\mathcal{Y}^{(i)}$ are either 0 or 1. For simplicity, we use $\widetilde{\mathcal{X}}^{(i,t)} \subseteq \mathcal{X}^{(i)}$ to denote the \textbf{feature history} containing all feature vectors until the $t$-th encounter, which means $ \widetilde{\mathcal{X}}^{(i,t)} = \{\bm x^{(i,1)}, \bm x^{(i,2)}, \cdots, \bm x^{(i,t)}\}$. The goal of prediction is using the  feature history $\widetilde{\mathcal{X}}^{(i,t)}$ and demographics $\bm d^{(i)}$
to predict the label $y^{(i,t)}$ for  the $(i)$-th patient
at the $t$-th encounter. For simplicity, we omit the patient index superscript $(i)$ unless it is explicitly needed.

\subsection{Structural Causal Model}

Figure \ref{fig:scm} shows the Structural Causal Model (SCM) \cite{pearl2010causal} describing one patient's EHRs.
A patient's EHR contains \textbf{demographics} $\bm D$ and multiple encounters (grey rectangles). We only plot the first and the $t$-th encounters in the figure for simplicity. Multiple encounters are arranged chronologically, and earlier encounters can affect later ones. We use a \textbf{recurrent updated latent factor} $\bm H$ to represent this mechanism. Within the $t$-th encounter, we can observe clinical \textbf{feature} $\bm X^t$ affects binary \textbf{label} $Y^t$. We also assume the presence of \textbf{unobserved confounders} $\bm Z^t$ that can simultaneously affect $\bm X^t$ and $Y^t$. In practice, latent factor $\bm Z^t$ can represent \textbf{underlying medical factors} such as patient's genotypes and lifestyle habits. Since demographics affect all encounters, it is connected with all $\bm X^t$ and  $Y^t$. Solid circles denote observed variables, and dotted circles denote latent variables.

\subsection{Sequential Counterfactual Fairness}
In our model, other fairness criteria can theoretically be utilized as well. However, in this paper, we specifically adopt Sequential Counterfactual Fairness  \cite{kusner2017counterfactual} to address health disparity in EHR because it has the same theoretical foundation (Pearl's causal inference theory \cite{pearl2010causal}) as our model.

Counterfactual probability refers to a probability based on a hypothetical scenario or an outcome that is contrary to what actually occurred. For example, $Y_{\bm D \leftarrow d}(\bm X)$ represents a counterfactual variable of $Y$ for a given feature $\bm X$ where the demographics $\bm D$ are replaced with $\bm D = d$, meaning ``What would the value of predicted label $Y$ had demographics $\bm D$ been set to a value $d$"? 
In this paper, we primarily aim to ensure that demographic features (such as gender and race) are treated fairly. According to the SCM in Figure \ref{fig:scm}, the counterfactual fairness of the $t$-th encounter is defined as:
\begin{align} \label{eq:cf_fair}
    & p(\hat{Y}_{\bm D \leftarrow \bm d}
    (\bm H^{t-1}, \bm Z^{t})=y|\widetilde{\bm X}^{t}=\widetilde{\mathcal{X}}^{t},\bm D=\bm d^{})
    =  \\ \nonumber
    & p(\hat{Y}_{\bm D \leftarrow \bm d'}(\bm H^{t-1},\bm Z^{t})=y|\widetilde{\bm X}^{t}=\widetilde{\mathcal{X}}^{t},\bm D=\bm d^{})
\end{align}

Here, $\hat{Y}$ is the predicted value of $Y$, $\widetilde{\bm X}^{t}$ denotes the \textbf{feature history} until encounter $t$ where $\widetilde{\mathcal{X}}^{t} = \{\bm x^1, \bm x^2, \cdots, \bm x^t\}$ is its realization. The fairness metric \ref{eq:cf_fair} means that the model should provide similar results for patients even with different demographics ($d$ and $d'$) under the same medical history. Here $d$ and $d'$ are two arbitrary  realizations of demographics $\bm D$.  As per the equation described, the quality and informativeness of the latent factors $\bm H^{t-1}$ and $\bm Z^{t}$ play a crucial role in maintaining counterfactual fairness. This is because the ultimate prediction $\hat{Y}$ is derived based on these latent factors.

%% file: Texts/2.2-Methods.tex
\begin{figure*}
    \centering
    \includegraphics[height=2.5in]{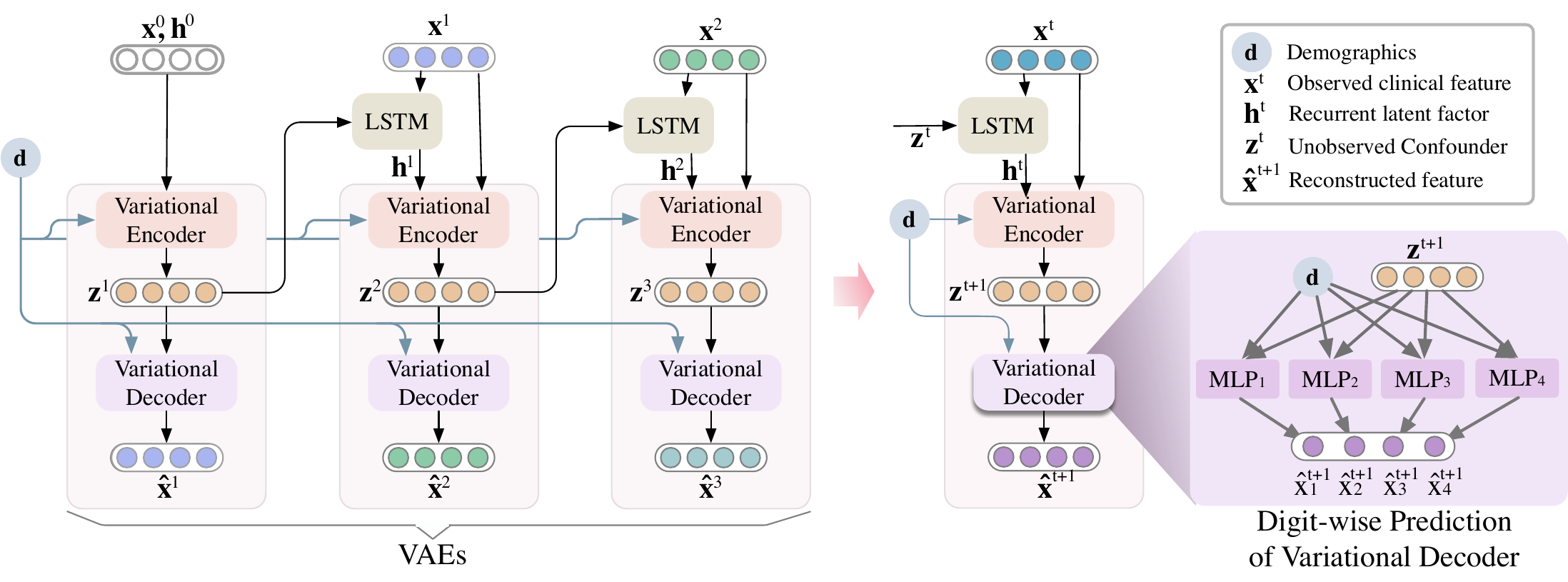}
    \caption{The architecture of the deep generative model used in
    FLMD's first learning phase.} 
    \label{fig:ovw}
\end{figure*}

\section{Method}
We propose a novel framework Fair Longitudinal Medical Deconfounder (FLMD) to derive fair and accurate predictions on EHR. FLMD contains two independent training stages: a deep generative model to capture the knowledge of unobserved confounders and an attentive prediction model for applying fairness constraints.

\subsection{Stage 1: Deep Generative Model}
The primary goal of this stage is to extract underlying medical factors from the observational features, which are essential for subsequent predictions and promoting fairness.
To achieve this, FLMD utilizes both the feature history $\widetilde{\mathcal{X}}^{t}$ until encounter $t$ and the demographics $\bm d$ as input to learn latent embeddings $\bm z^{t+1}$, $\bm h^{t+1}$ and predict the next encounter's feature $\bm x^{t+1}$ with them in a self-supervised manner. 
We integrate deep variational inference and recurrent modules to design this end-to-end model. Figure \ref{fig:ovw} demonstrates the model's architecture.

\paragraph{Variational Inference Module}
The Variational Inference Module aims to find the most proper latent factor $\bm z^{t+1}$ to reconstruct the next encounter feature $\bm x^{t+1}$, which means it should have estimated the posterior probability $p(\bm z^{t+1}|\bm x^{t+1}, \bm d)$.  However, due to the potential intractability of this probability and the demand of disentanglement between $\bm z^{t+1}$ and  $\bm d$, our model estimates a variational probability as a substitute. Similar to settings in \cite{kingma2013auto,chung2015recurrent,purushotham2017variational}, we assume $\bm z^{t+1}$ exhibits Gaussian distribution and 
use a neural network $\phi$ to determine its mean and variance. We have

\begin{align} \label{eq:reparameter1}
    [\bm \mu^{t+1}, \bm \sigma^{t+1}] & = \phi(\bm h^{t}, \bm x^t, \bm d), \\ 
    \bm z^{t+1} = & \bm \mu^{t+1} +\bm \sigma^{t+1} \odot \bm \epsilon
    \label{eq:reparameter2}
\end{align}

where $\odot$ denotes element-wise multiplication and $\bm \epsilon \sim \mathcal{N}(0, \bm I)$ is a random value with Gaussian distribution. Apart from the reliance shown in the SCM, we also add $\bm x^t$ as input to derive better results.

Later, the variational decoder takes the latent factor $\bm z^{t+1}$ as input to model the posterior $p(\bm x^{t+1}|{\bm z}^{t+1}, \bm d)$. Due to the Sequential Single Strong Ignorability \cite{bica2020time} that the deconfounder requires, we use independent predictive models to reconstruct each digit of $\bm x^{t+1}$ separately. Suppose the $j$-th feature of  $\bm x^{t+1}$ is denoted by  $x_j^{t+1}$, and then we use multiple independent neural networks

\begin{align} 
    \hat{x}_j^{t+1} = \chi_j(\bm z^{t+1}, \bm d)
\end{align}

to predict each digit of the feature vector.
where $\hat{x}_j^{t+1}$ is the reconstructed value of $x_j^{t+1}$ and $\chi_j(\cdot)$ is the model used to predict it. In this step,
every single feature owns an individual predictive model and the whole model for  $\bm x^{t+1}$ reconstruction is a set of models $\{\chi_1,\chi_2,\cdots,\chi_{|\bm x|}\}$. By conditioning on $\bm d$ in both the inference and reconstruction steps, the learned $\bm z^{t+1}$ can be marginally independent of $\bm d$ and contain information apart from demographics.

\paragraph{Recurrent Module}

According to the given SCM, the Recurrent Module aims to update $\bm h^{t+1}$ recurrently after obtaining $\bm z^{t+1}$ for the use of the next encounter.
In practice, we use a function $\bm h^{t+1} = \psi (\bm z^{t+1}, \bm x^{t+1})$ to achieve this goal. $\bm z^{t+1}$ is the variable connecting to $\bm h^{t+1}$ directly in SCM, and $\bm x^{t+1}$ is also contained as input of $\psi$ due to empirical reasons.

To initialize the whole sequential model, we first randomly initialize $\bm x^0$ and $\bm h^0$ as trainable parameter vectors. Then, the model utilizes them to sample $\bm z^1$ via Equation \ref{eq:reparameter1} and \ref{eq:reparameter2}. Finally, the $\bm h^1$ can be generated via function $\psi$ on the recurrent module.

The objective of this model $\mathcal{L}_1$ can be viewed as the reconstruction loss together with $\bm z$'s distribution penalty. Formally, the individual loss to predict the $i$-th patient's $(t+1)$-th encounter  is defined as
\begin{equation}
    \begin{aligned}
    & \mathcal{L}_1(\bm x^{(i,t)}, \bm h^{(i,{t})}, \bm d^{(i)}, \bm x^{(i,{t+1})}; \Theta_1) = 
    \\& - \text{KL}( q(\bm z^{(i,{t+1})}|\bm h^{(i,{t})}, \bm x^{(i,t)}, \bm d^{(i)}) ||  p(\bm z) )  
    \\& + \frac{1}{L} \sum_{l=1}^{L} \log p(\bm x^{(i,{t+1})}|\bm z^{(i,{t+1},l)},\bm d) + \lambda_1 ||\Theta_1||^2
    \end{aligned}
    \label{eq:loss}
\end{equation}

where $\Theta_1$ is the parameter set, $ q(\bm z^{t+1}|\bm h^{t}, \bm x^t, \bm d)$ and $p(\bm x^{t+1}|\bm z^{t+1},\bm d)$ are variational probability and the likelihood respectively,  $p(\bm z^{t+1}) \sim \mathcal{N}(0, \bm I)$ is  Gaussian prior and 
$L$ is the number of $z^{(i,{t})}$ samples. The overall objective is the average of the objective for each patient on each encounter.

\subsection{Fair Prediction with Counterfactual Objective} 

The learned $\bm z^{t}$ possesses three key characteristics: (1) it follows a Gaussian distribution, (2) it can effectively reconstruct next-encounter features, and (3) it is marginally independent of demographics. These properties enable it to yield relatively fair predictions. However, in order to further
mitigate health disparity and improve prediction accuracy, FLMD employs a separate predictive model and counterfactual loss in this stage.

The model in this stage accepts demographics $\bm d$, learned latent factor $\bm z^t$, and feature $\bm x^t$ to predict the label $y^t$. We adopt a Multi-head Attention Network \cite{vaswani2017attention} for prediction, and the loss function for single datapoint $(\bm d^{(i)}, \bm z^{(i,t)}, \bm x^{(i,t)} ,y^{(i,t)})$ can be denoted as
\begin{equation}
    \mathcal{L}_{F} = \sum_{i=1}^N \sum_{t=1}^{T{(i)}} \mathcal{J}(\text{Multi-head}(\bm d^{(i)}, \bm z^{(i,t)}, \bm x^{(i,t)}),y^{(i,t)})
\end{equation}
where $\text{Multi-head}(\bm d^{(i)}, \bm z^{(i,t)}, \bm x^{(i,t)})$ denotes the output of the Attention Network and $\mathcal{J}$ is the cross entropy loss. $N$ is the number of patients in the dataset and $T{(i)}$ is the number of encounters of patient $(i)$.

In order to maintain counterfactual fairness, we flip some sensitive attribute values in $\bm d$, obtain the counterfactual demographics $\bm d_{CF}$, and predict the counterfactual outcome with it. We hope the counterfactual prediction can still be the same as the original one, which means to minimize
\begin{equation}
    \mathcal{L}_{CF} =\sum_{i=1}^N \sum_{t=1}^{T{(i)}}  \mathcal{J}(\text{Multi-head}(\bm d_{CF}^{(i)}, \bm z^{(i,t)}, \bm x^{(i,t)})),y^{(i,t)}).
\end{equation}.

The loss of the prediction model in this stage $\mathcal{L}_2$ is the sum of the prediction loss and the counterfactual loss:
\begin{equation}
    \mathcal{L}_2 = \mathcal{L}_{F}+\lambda \mathcal{L}_{CF} + \lambda_2 ||\Theta_{2}||^2.
    \label{tloss}
\end{equation}
where $\lambda$ and $\lambda_2$ are hyperparameters and $\Theta_{2}$ denotes the parameters of models in this stage.

Moreover, more sophisticated approaches to guarantee counterfactual fairness such as counterfactual logit pairing \cite{garg2019counterfactual,pfohl2019counterfactual} can be used here. 

%% file: Texts/3-Exp.tex
\begin{table*}[]
\centering
\caption{Performance in terms of accuracy and fairness of all methods when the sensitive attribute is set as Race/Ethnicity on three prediction tasks: Prolonged Length of Stay (PLoS) (on Hip and Knee), Readmission (on Hip and Knee), and Procedure (on MIMIC-III). 
We highlight the best result in Bold and the second-best result with underlines.
}
\label{tab:q1t1}
\begin{tabular}{l|llll|llll|ll}
\hline \hline
Dataset        & \multicolumn{4}{c|}{Hip}                                                                                            & \multicolumn{4}{c|}{Knee}                                                                                           & \multicolumn{2}{c}{MIMIC-III}                           \\ \hline
Task          & \multicolumn{2}{c|}{PLoS}                                & \multicolumn{2}{c|}{Readmission}                         & \multicolumn{2}{c|}{PLoS}                                & \multicolumn{2}{c|}{Readmission}                         & \multicolumn{2}{c}{Procedure}                           \\ \hline
Metrics        & \multicolumn{1}{c}{AUC} & \multicolumn{1}{c|}{$\Delta_{HD}^{Binary}$} & \multicolumn{1}{c}{AUC} & \multicolumn{1}{c|}{$\Delta_{HD}^{Binary}$} & \multicolumn{1}{c}{AUC} & \multicolumn{1}{c|}{$\Delta_{HD}^{Binary}$} & \multicolumn{1}{c}{AUC} & \multicolumn{1}{c|}{$\Delta_{HD}^{Binary}$} & \multicolumn{1}{c}{$nDCG_5$} & \multicolumn{1}{c}{$\Delta_{HD}^{Multi}$} \\ \hline
Transformer    & \underline{0.7238}            & \multicolumn{1}{l|}{1.492}          & 0.6373                  & 1.751                          & \underline{0.6886}            & \multicolumn{1}{l|}{1.411}         & \underline{0.6608}            & 1.757                          & \textbf{0.5883}         & 8.105                         \\
CF Prediction  & 0.7129                  & \multicolumn{1}{l|}{1.354}          & 0.625                   & 1.629                          & 0.6772                  & \multicolumn{1}{l|}{1.294}         & 0.6323                  & 1.658                          & 0.5731                  & 6.596                         \\
Med Deconf     & 0.7187                  & \multicolumn{1}{l|}{1.3}            & 0.6324                  & 1.579                          & 0.6846                  & \multicolumn{1}{l|}{1.282}         & 0.6467                  & 1.637                          & 0.5806                  & 6.326                         \\
CRN            & 0.7217                  & \multicolumn{1}{l|}{1.341}          & 0.6319                  & 1.613                          & 0.6866                  & \multicolumn{1}{l|}{1.326}         & 0.6488                  & 1.685                          & 0.5855                  & 6.578                         \\
TE-CDE         & 0.7232                  & \multicolumn{1}{l|}{1.297}          & 0.633                   & 1.518                          & 0.6874                  & \multicolumn{1}{l|}{1.272}         & 0.6529                  & 1.631                          & 0.5869                  & 6.394                         \\ \hline
FLMD           & \textbf{0.7254}         & \multicolumn{1}{l|}{\textbf{1.273}} & \textbf{0.6401}         & \textbf{1.465}                 & \textbf{0.6889}         & \multicolumn{1}{l|}{\textbf{1.25}} & \textbf{0.6619}         & \textbf{1.575}                 & \underline{0.5877}            & \textbf{6.149}                \\
FLMD + Eq odds & 0.7231                  & \multicolumn{1}{l|}{\underline{1.289}}    & \underline{0.6398}            & \underline{1.484}                    & 0.6882                  & \multicolumn{1}{l|}{\underline{1.257}}   & 0.6545                  & \underline{1.611}                    & 0.5859                  & \underline{6.232}                   \\
IPW + FLMD     & 0.7214                  & \multicolumn{1}{l|}{1.318}          & 0.6326                  & 1.558                          & 0.6862                  & \multicolumn{1}{l|}{1.312}         & 0.6464                  & 1.714                          & 0.5772                  & 6.988                         \\ \hline \hline
\end{tabular}
\end{table*}


\begin{table*}[]
\centering
\caption{Performance in terms of accuracy and fairness of all methods when the sensitive attribute is set as Gender on three prediction tasks: Prolonged Length of Stay (PLoS) (on Hip and Knee), Readmission (on Hip and Knee), and Procedure (on MIMIC-III). 
We highlight the best result in Bold and the second-best result with underlines.}
\label{tab:q1t2}
\begin{tabular}{l|llll|llll|ll}
\hline \hline
Dataset        & \multicolumn{4}{c|}{Hip}                                                                                            & \multicolumn{4}{c|}{Knee}                                                                                           & \multicolumn{2}{c}{MIMIC-III}                           \\ \hline
Task          & \multicolumn{2}{c|}{PLoS}                                & \multicolumn{2}{c|}{Readmission}                         & \multicolumn{2}{c|}{PLoS}                                & \multicolumn{2}{c|}{Readmission}                         & \multicolumn{2}{c}{Procedure}                           \\ \hline
Metrics        & \multicolumn{1}{c}{AUC} & \multicolumn{1}{c|}{$\Delta_{HD}^{Binary}$} & \multicolumn{1}{c}{AUC} & \multicolumn{1}{c|}{$\Delta_{HD}^{Binary}$} & \multicolumn{1}{c}{AUC} & \multicolumn{1}{c|}{$\Delta_{HD}^{Binary}$} & \multicolumn{1}{c}{AUC} & \multicolumn{1}{c|}{$\Delta_{HD}^{Binary}$} & \multicolumn{1}{c}{$nDCG_5$} & \multicolumn{1}{c}{$\Delta_{HD}^{Multi}$} \\ \hline
Transformer    & 0.7238                  & \multicolumn{1}{l|}{1.239}          & 0.6373                  & 1.606                          & 0.6886                  & \multicolumn{1}{l|}{1.044}          & 0.6608                  & 1.521                          & \textbf{0.5883}         & 3.772                         \\
CF Prediction  & 0.7129                  & \multicolumn{1}{l|}{1.160}          & 0.625                   & 1.455                          & 0.6772                  & \multicolumn{1}{l|}{0.974}          & 0.6323                  & 1.321                          & 0.5731                  & 3.218                         \\
Med Deconf     & 0.7229                  & \multicolumn{1}{l|}{1.003}          & 0.6216                  & 1.394                          & 0.6852                  & \multicolumn{1}{l|}{\underline{0.946}}    & 0.6348                  & 1.229                          & 0.5786                  & 3.055                         \\
CRN            & 0.7254                  & \multicolumn{1}{l|}{1.184}          & 0.6241                  & 1.495                          & 0.6885                  & \multicolumn{1}{l|}{0.982}          & 0.6457                  & 1.391                          & 0.5821                  & 3.111                         \\
TE-CDE         & \underline{0.7263}            & \multicolumn{1}{l|}{1.082}          & 0.6339                  & 1.389                          & 0.6898                  & \multicolumn{1}{l|}{0.956}          & 0.6556                  & 1.189                          & 0.5848                  & \underline{2.983}                   \\ \hline
FLMD           & \textbf{0.7265}         & \multicolumn{1}{l|}{\textbf{0.955}} & \textbf{0.6457}         & \textbf{1.348}                 & \textbf{0.6913}         & \multicolumn{1}{l|}{\textbf{0.924}} & \textbf{0.6681}         & \textbf{1.095}                 & \textbf{0.5901}         & \textbf{2.798}                \\
FLMD + Eq odds & 0.7248                  & \multicolumn{1}{l|}{\underline{0.9810}}    & \underline{0.6424}            & \underline{1.377}                    & \underline{0.691}             & \multicolumn{1}{l|}{0.968}          & \underline{0.6643}            & \underline{1.126}                    & 0.5874                  & 3.063                         \\ 
IPW + FLMD     & 0.7214                  & \multicolumn{1}{l|}{1.050}          & 0.6326                  & 1.471                          & 0.6862                  & \multicolumn{1}{l|}{0.993}          & 0.6464                  & 1.366                          & 0.5772                  & 3.321                         \\
\hline \hline
\end{tabular}
\end{table*}


\begin{figure*}[t]
    \begin{subfigure}[]{0.23\textwidth}
        \centering
\includegraphics[width=\linewidth,trim=0 19 57 75,,clip]{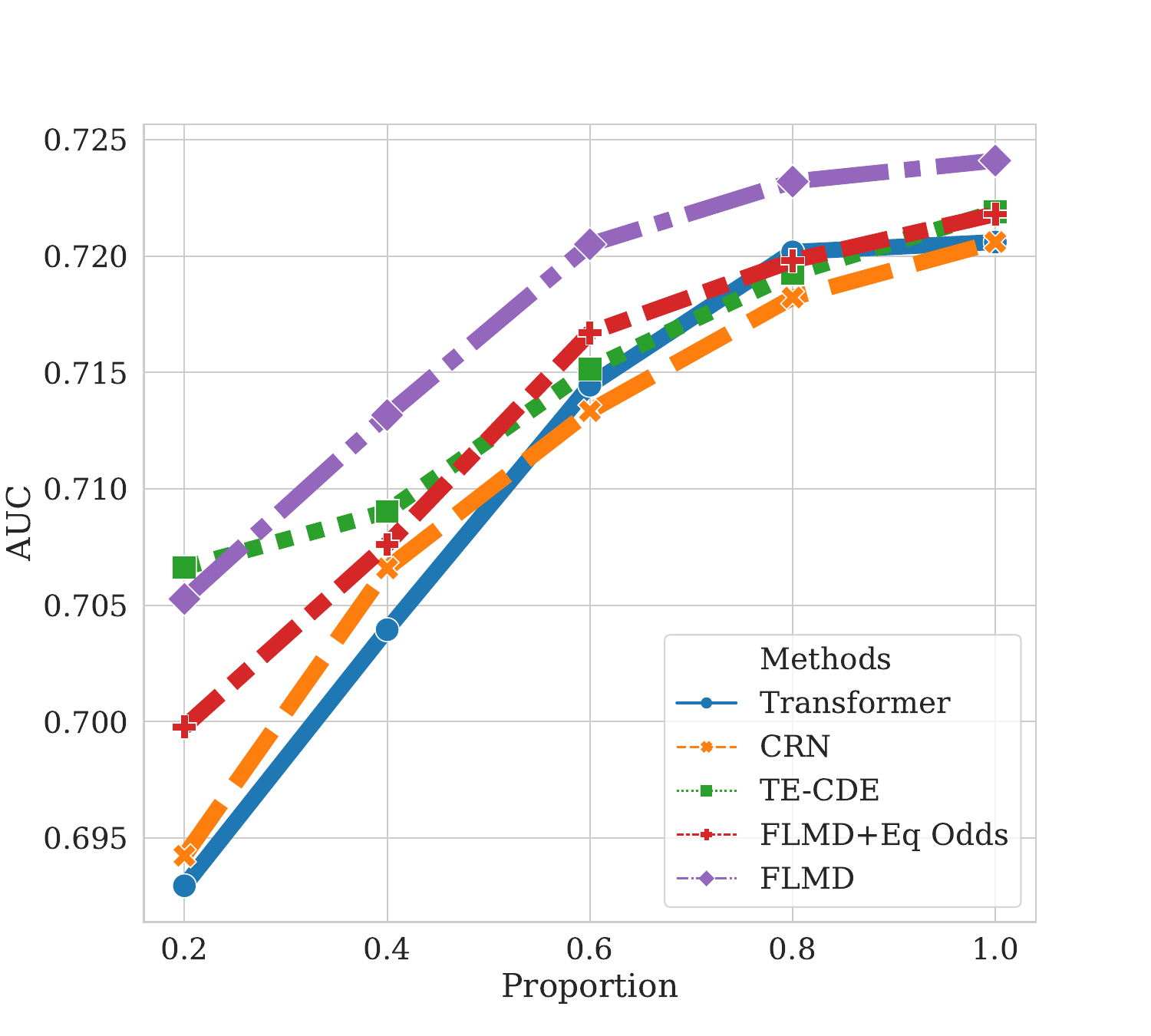}
        \caption{PLoS on Hip.
        } 
        \label{fig:11}
    \end{subfigure}
    \begin{subfigure}[]{0.23\textwidth}
        \centering
\includegraphics[width=\linewidth,trim=0 19 57 75,,clip]{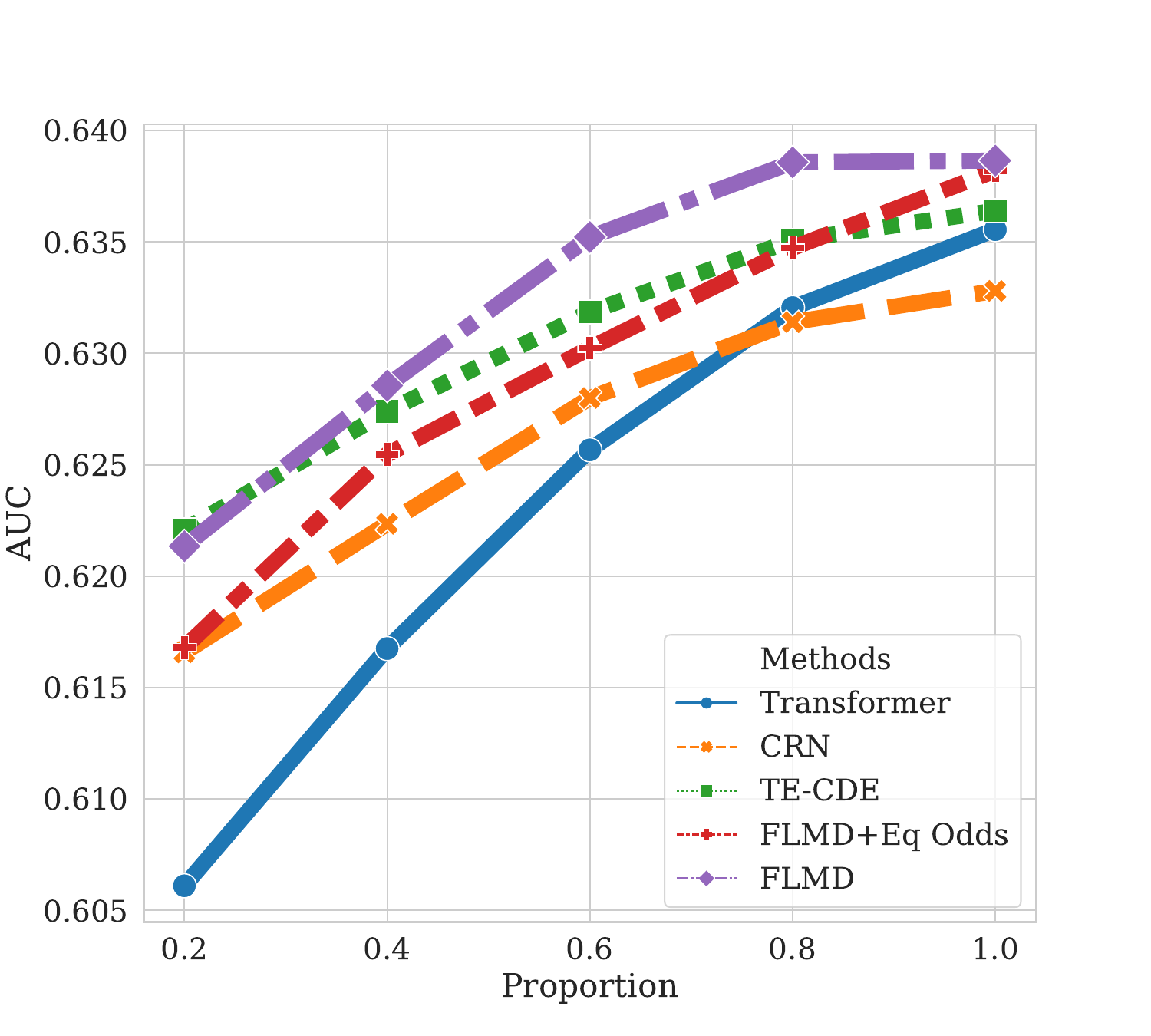}
        \caption{Readmission on Hip.
        } 
        \label{fig:12}
    \end{subfigure}
    \begin{subfigure}[]{0.23\textwidth} 
        \centering
\includegraphics[width=\linewidth,trim=0 19 57 75,,clip]{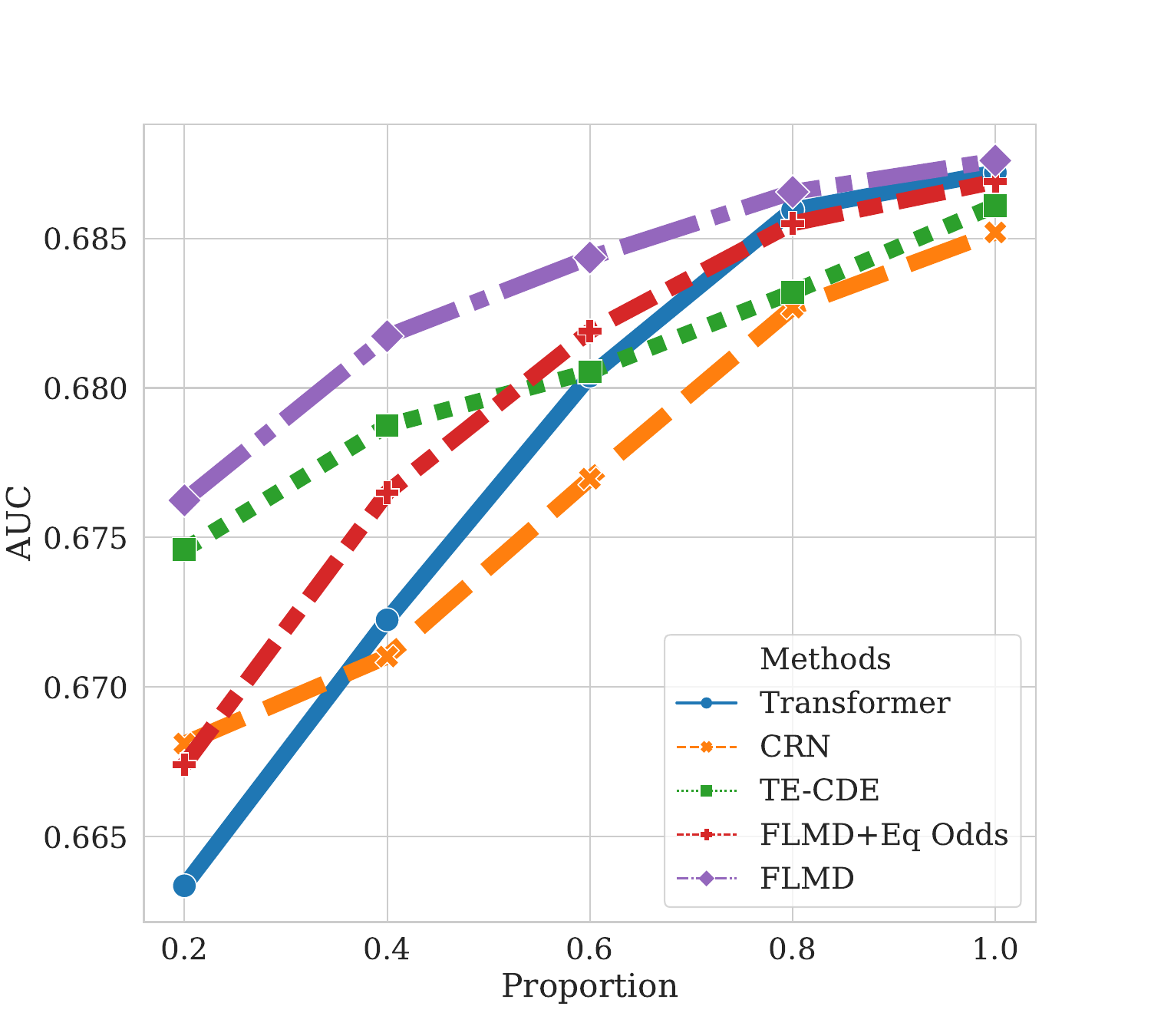}
        \caption{PLoS on Knee.
        }
       \label{fig:13}
    \end{subfigure}
    \begin{subfigure}[]{0.23\textwidth} 
        \centering
\includegraphics[width=\linewidth,trim=0 19 57 75,,clip]{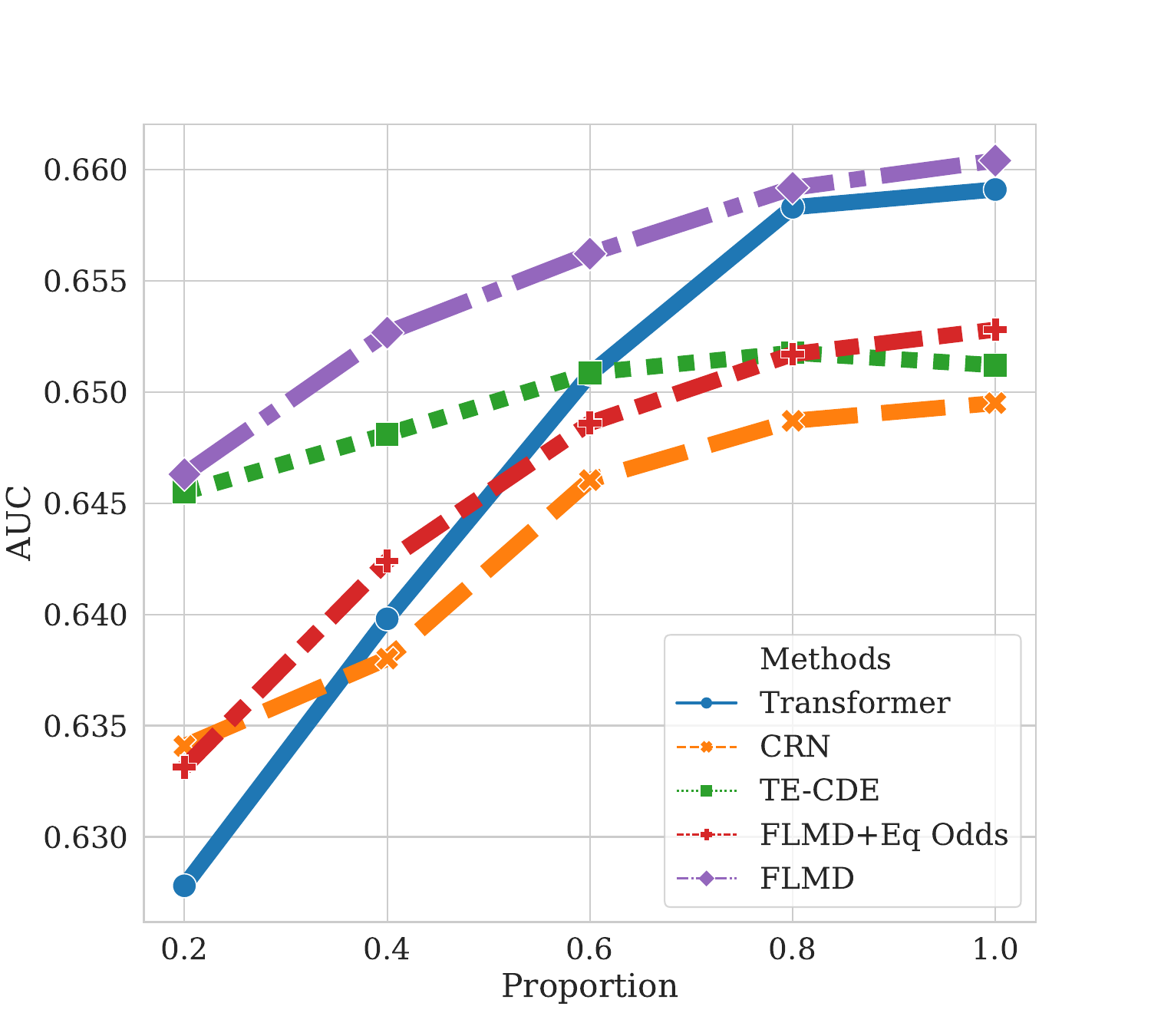}
        \caption{Readmission on Knee.
        }
       \label{fig:14}
    \end{subfigure}

\vspace{2mm} 

    \begin{subfigure}[]{0.23\textwidth}
        \centering
\includegraphics[width=\linewidth,trim=0 19 57 75,,clip]{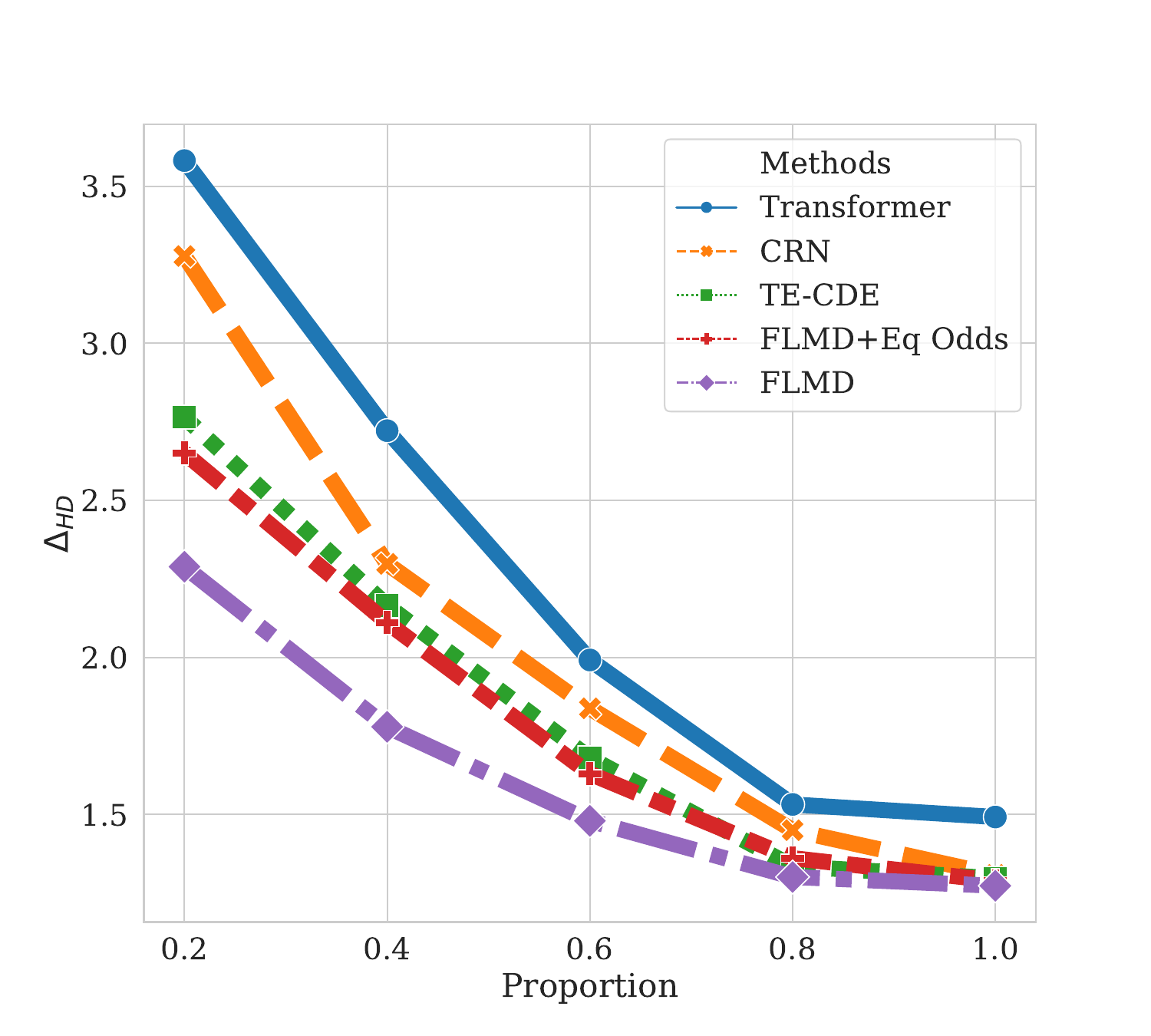}
        \caption{PLoS on Hip.
        } 
        \label{fig:21}
    \end{subfigure}
    \begin{subfigure}[]{0.23\textwidth}
        \centering
\includegraphics[width=\linewidth,trim=0 19 57 75,,clip]{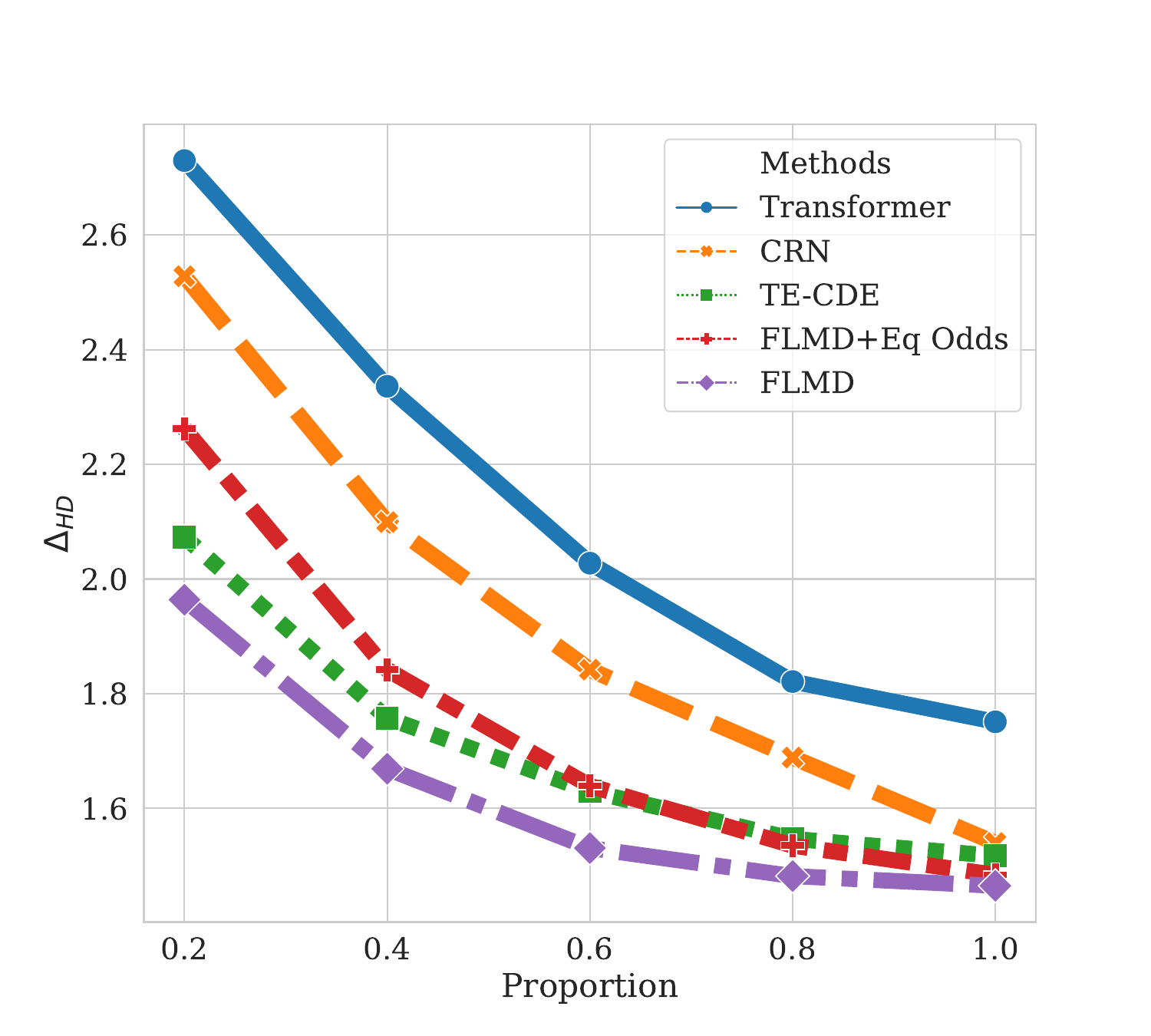}
        \caption{Readmission on Hip.
        } 
        \label{fig:22}
    \end{subfigure}
    \begin{subfigure}[]{0.23\textwidth} 
        \centering
\includegraphics[width=\linewidth,trim=0 19 57 75,,clip]{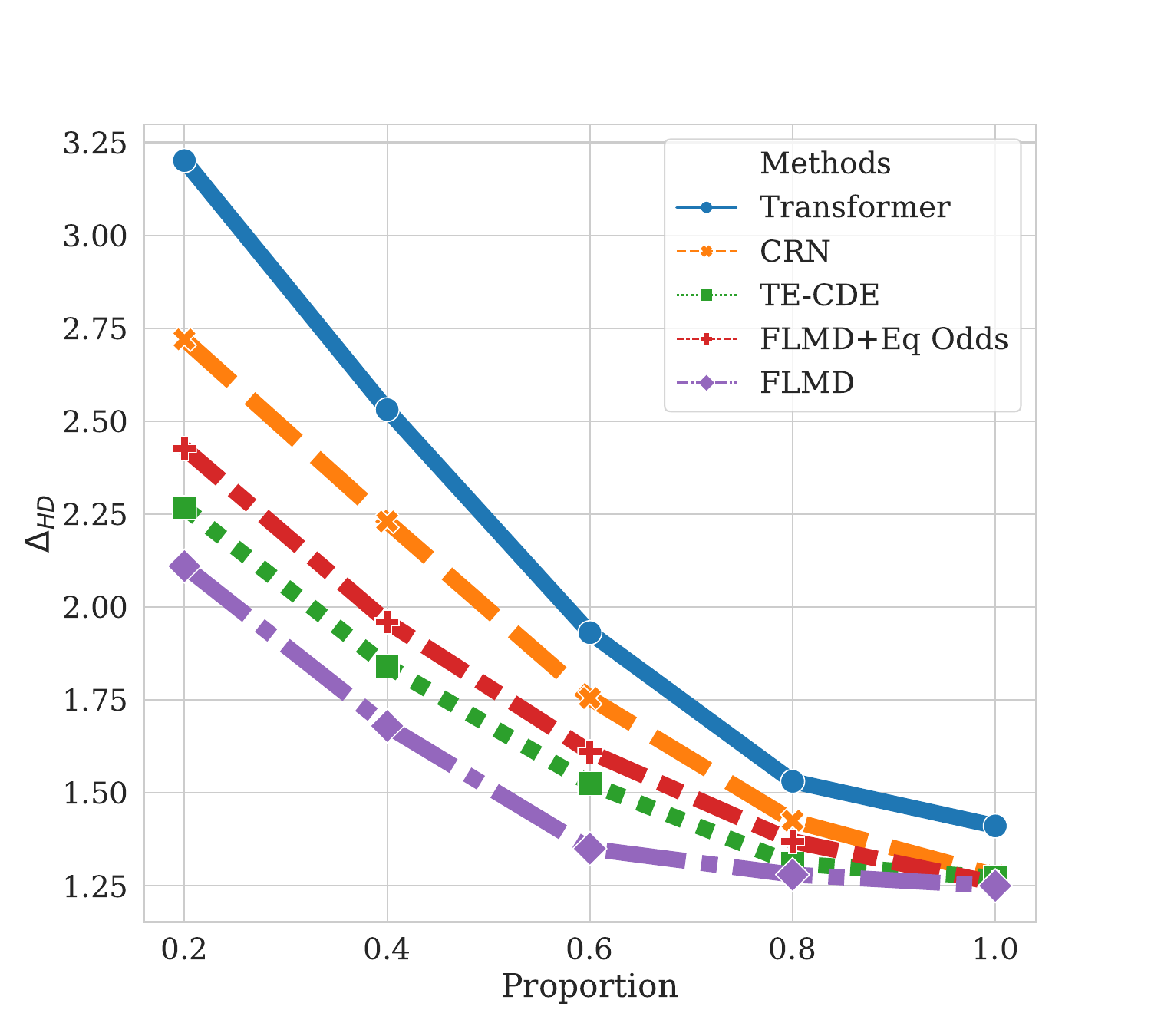}
        \caption{PLoS on Knee.
        }
        \label{fig:23}
    \end{subfigure}
    \begin{subfigure}[]{0.23\textwidth} 
        \centering
\includegraphics[width=\linewidth,trim=0 19 57 75,,clip]{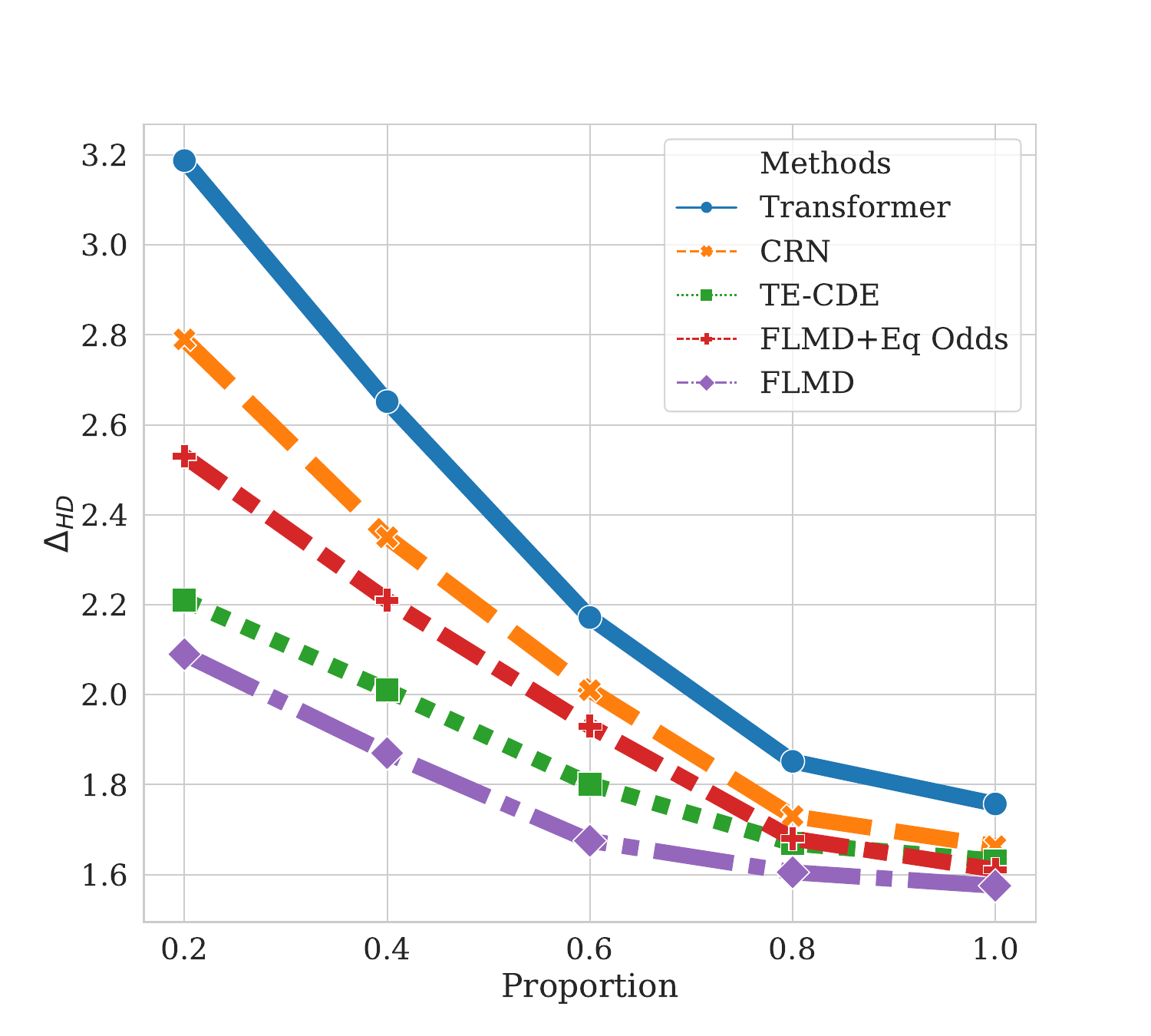}
        \caption{Readmission on Knee.
        }
        \label{fig:24}
    \end{subfigure}

\caption{Line plots in terms of model accuracy and fairness on disturbed data. The horizontal axis represents the proportion of original (undisturbed) data used for training, and the vertical axis represents model accuracy/fairness. Figures (a)-(d) illustrate the changes in model accuracy (AUC) on specific tasks within particular datasets, while Figures (e)-(h) visualize the changes ($\Delta_{HD}^{Binary}$) in fairness on those same tasks within the corresponding datasets.
}  
\label{tab:q2}
\end{figure*}


\section{Experiments}

In this section, we conduct extensive experiments to show the effectiveness of the proposed model. Specifically, we aim to answer the following research questions:
\begin{itemize} 
    \item How does the performance of FLMD compare to other fairness methods in terms of fairness and accuracy when applied to Electronic Health Records (EHRs)? Specifically, which variant of FLMD demonstrates the highest level of performance?
    (\textbf{Q1})? 
    \item Does FLMD exhibit superior generalization capabilities compared to other models? Particularly, when trained on biased (disturbed or imbalanced) data, does FLMD retain its ability to uphold fairness and accuracy guarantees?
    (\textbf{Q2})? 
    \item Does the latent embedding derived from FLMD's first stage possess sufficient information to capture the underlying factors (unobserved confounders) effectively?
    (\textbf{Q3})?
\end{itemize}

\subsection{Datasets}

We use two real-world EHR datasets to measure FaStMeD and other methods' performances.

\begin{itemize}
    
    \item \textbf{\textit{MIMIC-III\footnote{https://physionet.org/content/mimiciii/1.4/}}}. MIMIC-III \cite{johnson2016mimic} is a deidentified, publicly-available dataset comprising comprehensive clinical data of patients admitted to the Beth Israel Deaconess Medical Center. MIMIC-III contains EHRs associated with 46,520 patients including over 20 tables, such as medical events, diagnoses, prescriptions, etc.
    
    \item \textbf{\textit{NSQIP\footnote{https://www.facs.org/quality-programs/data-and-registries/acs-nsqip/}}}. 
    The National Surgical Quality Improvement Program (NSQIP) \cite{cohen2016improved} collects patient records associated with surgical operations, including information both before and after the procedure. This paper primarily focuses on two subsets of the NSQIP dataset, referred to as ``\textbf{Hip}" and ``\textbf{Knee}," which pertain to patients undergoing Hip Arthroplasty  (CPT 27130)  and Total Knee Arthroplasty (CPT 27447) operations, respectively.
    The numbers of patients in these two subsets are 96,441 and 156,292. 

\end{itemize}

\paragraph{\textbf{Experiment Settings}}
In MIMIC-III, one patient may have multiple encounters. We use lab measurements, diagnoses, and clinical events for each encounter as model input to predict procedures from the 157 most common ICD codes. We regard the patient's ethnicity and gender as sensitive attributes. 
In NSQIP experiments, we use pre-operation features (including smoke, dyspnea, COPD, steroid usage, bleeding disorder, etc.) to predict the Prolonged Length of Stay (PLoS) and Readmission. We take gender and race as sensitive attributes. All sensitive attributes are preprocessed into binary features.
For ethnicity/race, patients are represented as  White/non-White; for sex, patients are represented as male/female. The ratio of train:validation:test data is 7:2:1.

\subsection{Implementation Details}
In the first training stage, we set function $\psi$ to be LSTM \cite{hochreiter1997long} while other recurrent units such as GRU or Transformer can also be used here.
We set $\phi$ and $\chi_j(\cdot)$ to be individual Multi-layer Perceptrons (MLPs) and their capacities are set to be 512 and 16 respectively.

Since the first stage of FLMD is unsupervised, we train a model on the whole dataset and get the latent vector $\bm z$ for all patients and all encounters. In this step, we set up the learning rate as $10^{-5}$ and the weight decay as $10^{-7}$.
In the second phase, we use the pre-trained latent vectors $\bm z$ obtained in the first phase together with patient features and factual/counterfactual demographics to train a supervised model on the training set. We randomly flip the value of sensitive attributes and include the counterfactual samples as part of the training set.
We also use a validation set to tune the hyper-parameters and a test set to evaluate the model's performance. The ratio of training, validation, and test sets is 7:2:1. In this step, we set up the length of latent vectors $\bm z$ as 256,
the learning rate as $10^{-5}$ and the weight decay varies for different datasets.

\subsection{Evaluation Metrics}

\paragraph{Accuracy Metrics} 
We use the Area Under the Receiver Operating Characteristic curve ($AUC$) to measure binary classifications' accuracy. We use Normalized Discounted Cumulative Gain at five ($nDCG_5$) to measure the accuracy of multi-class classifications. 

\paragraph{Fairness Metrics} To measure the extent of health disparity, we calculate the accuracy difference between demographic subgroups. Suppose $\bm G_1$ and $\bm G_2$ are two patient subgroups with different sensitive attributes (i.e., White and non-White patients). Then the health disparity means the difference of a model's accuracy between them, which is defined as
\begin{equation}
    \Delta^{Binary}_{HD} = | AUC^{\bm G_1}_5 - AUC^{\bm G_2}_5 | \times 10^3
\end{equation}
for binary classification and

\begin{equation}
    \Delta^{Multi}_{HD} = | nDCG^{\bm G_1}_5-nDCG^{\bm G_2}_5 | \times 10^3
\end{equation}
for multi-class classification. Here, we multiply the difference of accuracy by 1000, as the original difference is too small to evaluate effectively.

\subsection{Comparison Experiments}

We compare our FLMD  with the following models and fairness methods to measure their performance in terms of accuracy and fairness. 

\begin{itemize}
    \item \textbf{Transformer }\cite{vaswani2017attention}. We adopt multi-headed attention layers in Transformer to model EHR. This method doesn't address any fairness issue but is presented as a baseline.

    \item \textbf{CF Prediction} \cite{pfohl2019counterfactual}.
    Pfohl et al. propose a method based on Causal Effect VAE \cite{louizos2017causal} to make counterfactual predictions for clinical data. 
    
    \item \textbf{Med Deconf}\cite{zhang2019medical}.
    The Medical Deconfounder is another application of the deconfounder theory  \cite{wang2019blessings,wang2020towards}. We use the same setting as the original paper introduced to derive results.

    \item  \textbf{CRN}. \cite{bica2020estimating} Counterfactual Recurrent Network is a seq-to-seq model using domain adversarial training to  answer counterfactual medical questions.

    \item \textbf{TE-CDE}. \cite{seedat2022continuous} TE-CDE is another counterfactual sequential model 
    to answer counterfactual medical questions.
 
\end{itemize}

Apart from the above fairness methods, we also compare FLMD with the following variants:
\begin{itemize}
    \item \textbf{FLMD + Eq odds} 
    Eq odds\cite{hardt2016equality} is a fairness paradigm pursuing the same equal false positive rate for all groups. We adopt the post-processing methods presented in \cite{romano2020achieving} to ensure this fairness.
    
    \item \textbf{IPW + FLMD} \cite{imbens2015causal}. Inverse Propensity Weighting is a traditional method to derive unbiased estimation for observational data. We adopt IPW as a pre-processing method to mitigate health disparity.

\end{itemize}


\input{Texts/3.1-Q1}

\input{Texts/3.2-Q2}

\input{Texts/3.3-Q3}
\input{Texts/3.4-Q4}

%% file: Texts/3.1-Q1.tex
\subsection{Analysis of Accuracy and Fairness (\textbf{Q}1)}

The performance of all methods in terms of accuracy and fairness on EHR prediction is presented in Table \ref{tab:q1t1} and \ref{tab:q1t2}. To obtain these tables, we conduct experiments on three datasets (Hip, Knee and MIMIC-III) with three tasks: Prolonged Length of Stay (PLoS), Readmission, and Procedure prediction. 
We highlight the best result in Bold and the second-best result with underlines. 
In particular, certain methods include adjustable hyper-parameters to accommodate varying levels of fairness-accuracy trade-offs. In these tables, we present the results corresponding to the parameter settings that maximize the multiplication of fairness (i.e., $1-\Delta_{HD}$) and accuracy (i.e, AUC or $nDCG_5$).

\paragraph{Accuracy Analysis}
In Table \ref{tab:q1t1} and \ref{tab:q1t2}, the columns under AUC and $nDCG_5$ reveal the accuracy of all methods on different tasks. Our proposed FLMD equipped with the counterfactual fairness objective, achieves the highest accuracy across all baselines on almost all experiments.
The Transformer model stands out as the second most accurate method. However, it solely focuses on optimizing accuracy and does not specifically address fairness concerns in predictions.
Remarkably, FLMD demonstrates superior performance over the Transformer in the majority of experiments, thus showcasing the effectiveness of the FLMD architecture.  In addition to these methods, FLMD + EQ Odds, TE-CDE, and CRN exhibit lower prediction accuracy when compared to FLMD and Transformer, but they still offer relatively high-quality predictions.

\paragraph{Fairness Analysis}
The columns under $\Delta_{HD}^{Binary}$ and $\Delta_{HD}^{Multi}$ provide insights into the fairness of all methods. Lower values indicate the model's greater fairness. According to the tables, we can observe FLMD also achieves the lowest health disparity compared with all baselines on all experiments. 
This highlights FLMD's capability to generate fair outcomes. FLMD + Eq Odds achieves the second-highest level of fairness across all tasks. In addition to them, IPW + FLMD, TE-CDE, and Med Deconf can also produce relatively fair predictions. The Transformer produces the least fair results in all experiments, as its primary focus is on optimizing accuracy rather than fairness.

In summary, the experiment results support the existence of an accuracy/fairness trade-off. If a model aims to achieve fairer results, it is more likely to experience a decrease in prediction accuracy. However, our proposed FLMD can mitigate this effect and achieve fairness and relatively high accuracy simultaneously. 
We attribute FLMD's superior performance of to two main factors: its architecture in modeling latent factors and the adoption of a counterfactual fairness objective.
FLMD's architecture allows it to effectively model latent factors, enabling the model to learn underlying knowledge from the data. 
The counterfactual fairness objective enables the adjustment of parameters within the model itself. Therefore, this in-processing fairness approach allows for a more generalizable model compared to other approaches.
In addition to FLMD, FLMD + Eq Odds and TE-CDE achieve second-tier results. FLMD + Eq Odds shares the same model architecture as FLMD, while TE-CDE employs the same counterfactual fairness paradigm as FLMD. The consistent performance of these methods reinforces the effectiveness of the chosen model architecture and the incorporation of a counterfactual fairness objective in achieving fairer outcomes.

%% file: Texts/3.2-Q2.tex
\begin{figure*}[t]
    \begin{subfigure}[]{0.23\textwidth}
        \centering
\includegraphics[width=\linewidth,trim=8 13 57 51,,clip]{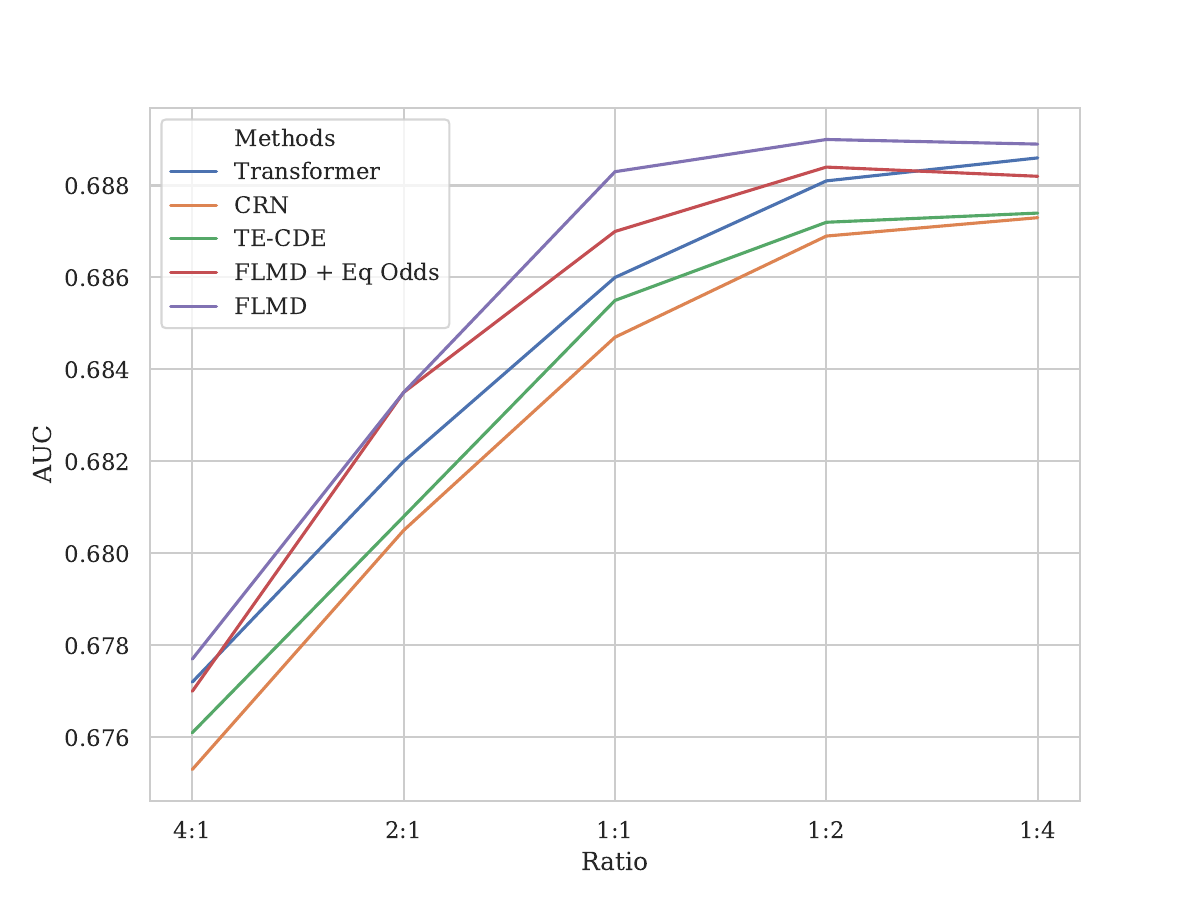}
        \caption{} 
        \label{fig:31}
    \end{subfigure}
    \begin{subfigure}[]{0.23\textwidth}
        \centering
\includegraphics[width=\linewidth,trim=8 13 57 51,,clip]{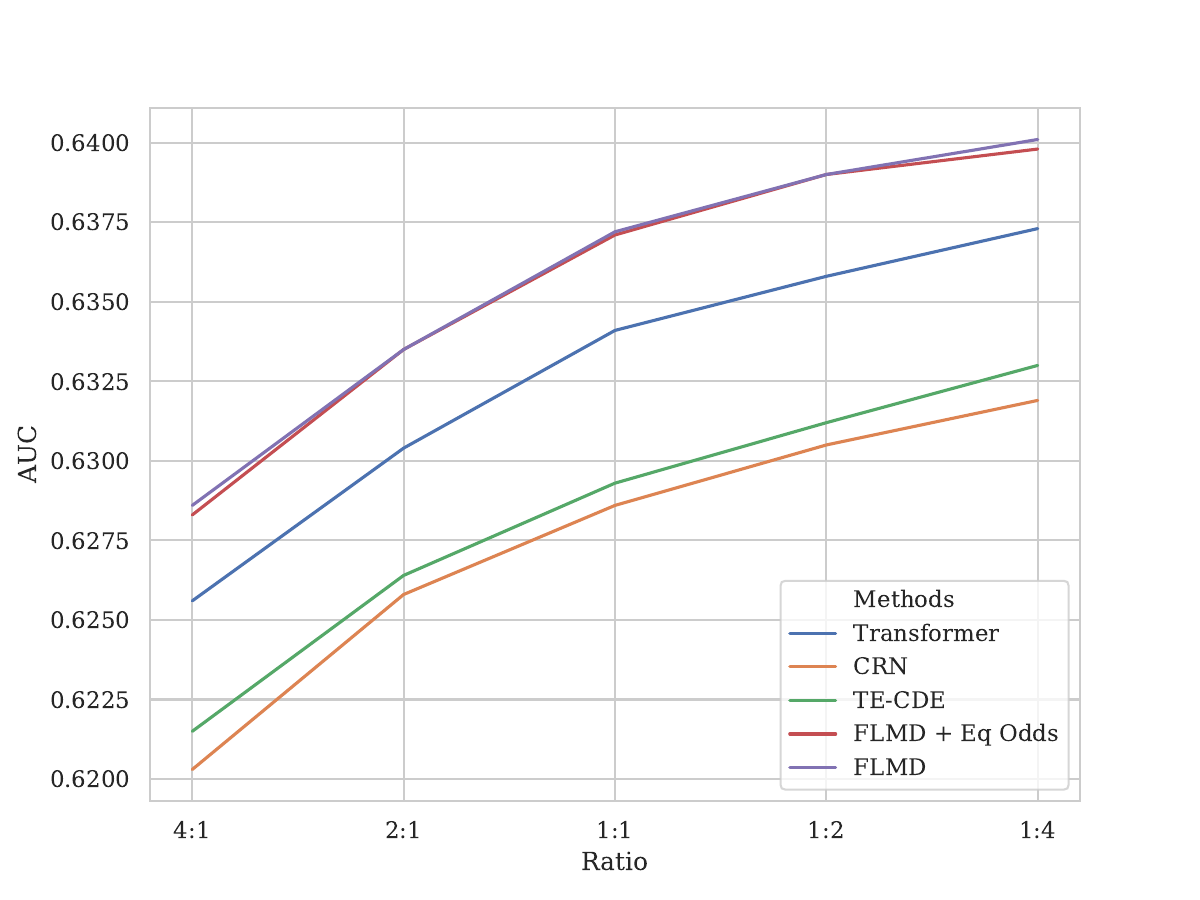}
        \caption{} 
        \label{fig:32}
    \end{subfigure}
    \begin{subfigure}[]{0.23\textwidth}
        \centering
\includegraphics[width=\linewidth,trim=8 13 57 51,,clip]{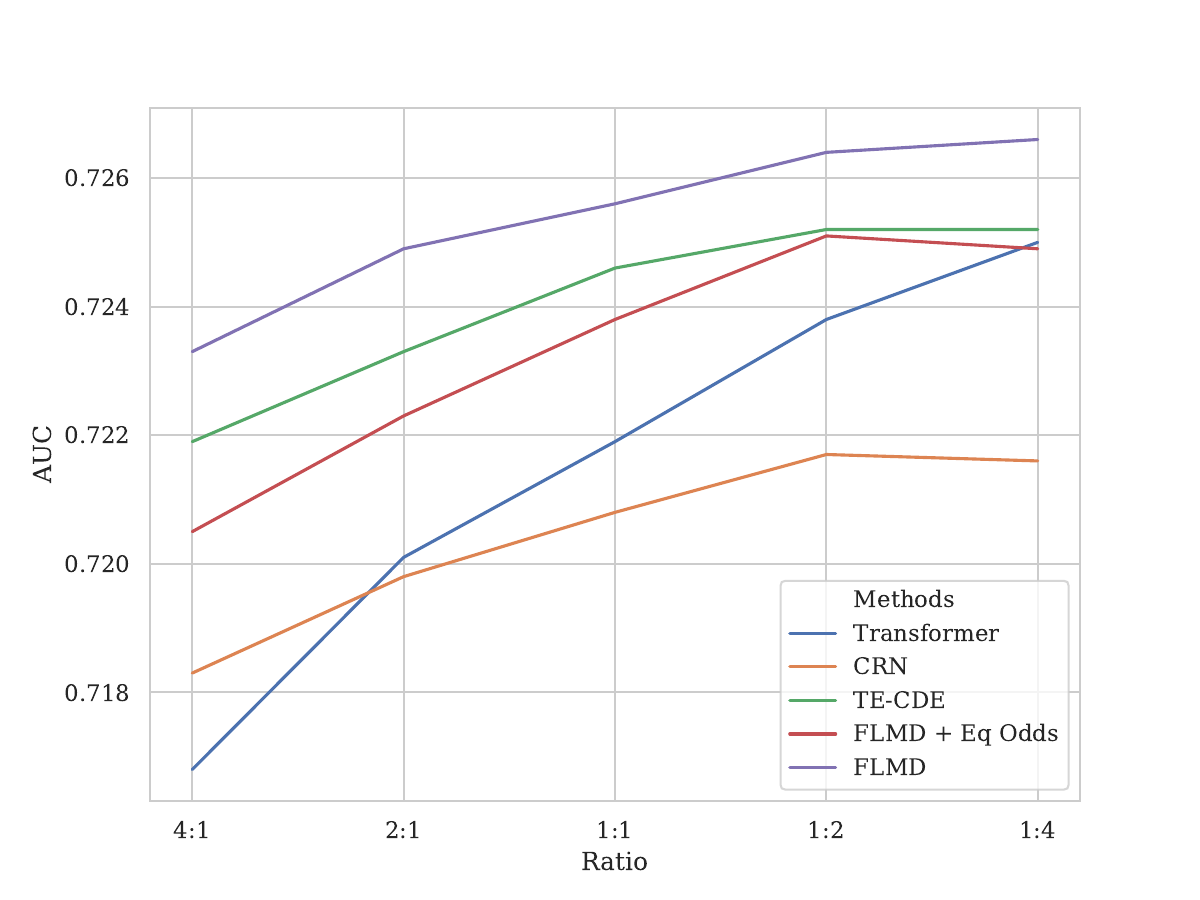}
        \caption{} 
        \label{fig:33}
    \end{subfigure}
    \begin{subfigure}[]{0.23\textwidth}
        \centering
\includegraphics[width=\linewidth,trim=8 13 57 51,,clip]{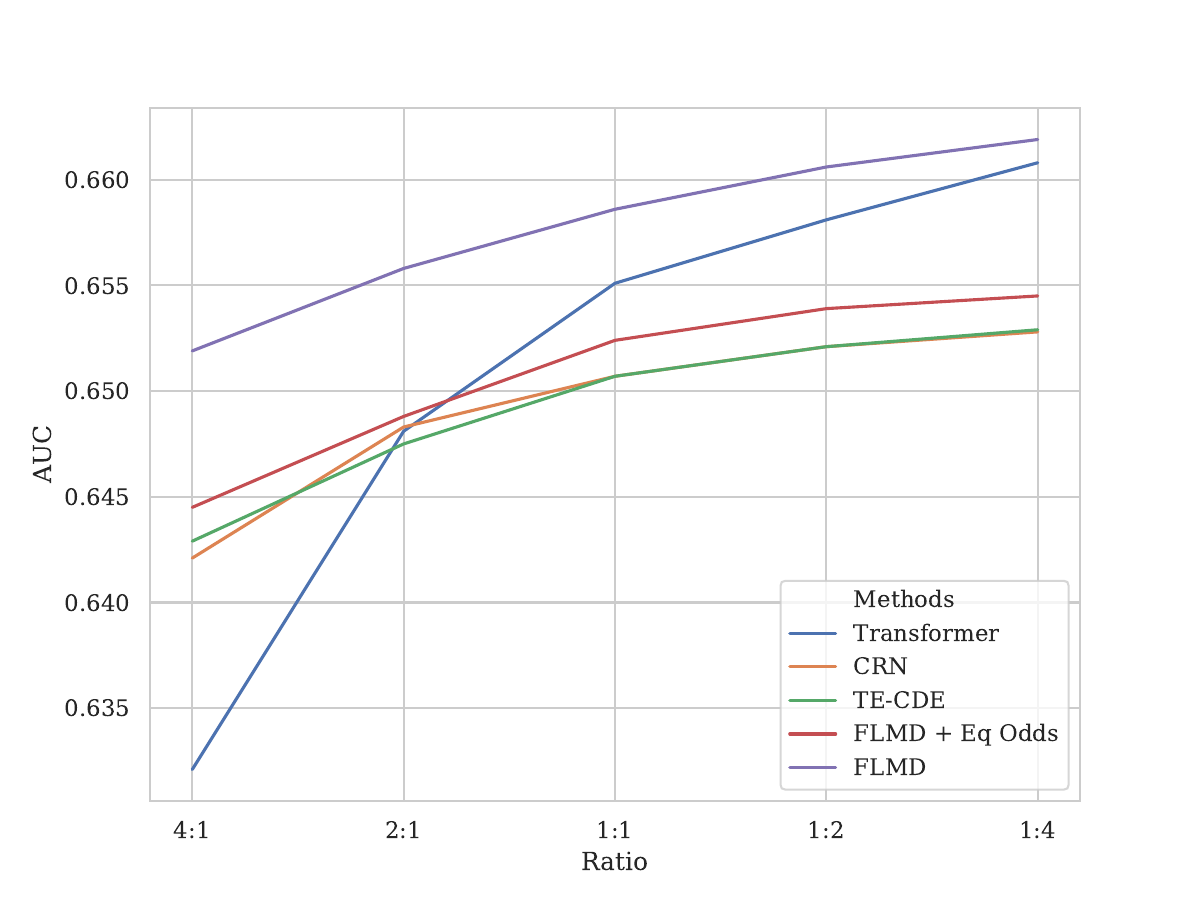}
        \caption{} 
        \label{fig:34}
    \end{subfigure}

\vspace{5pt}

    \begin{subfigure}[]{0.23\textwidth}
        \centering
\includegraphics[width=\linewidth,trim=8 13 57 51,,clip]{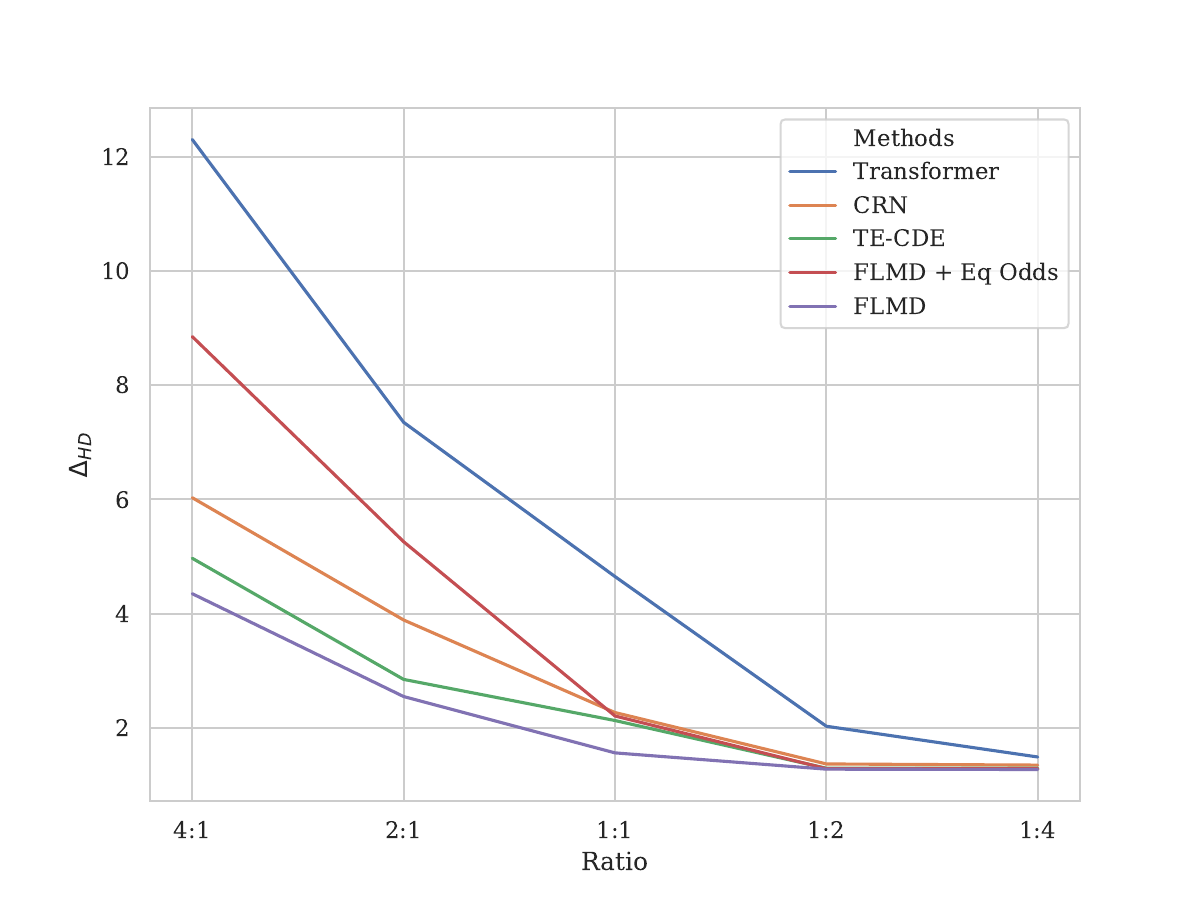}
        \caption{} 
        \label{fig:35}
    \end{subfigure}
    \begin{subfigure}[]{0.23\textwidth}
        \centering
\includegraphics[width=\linewidth,trim=8 13 57 51,,clip]{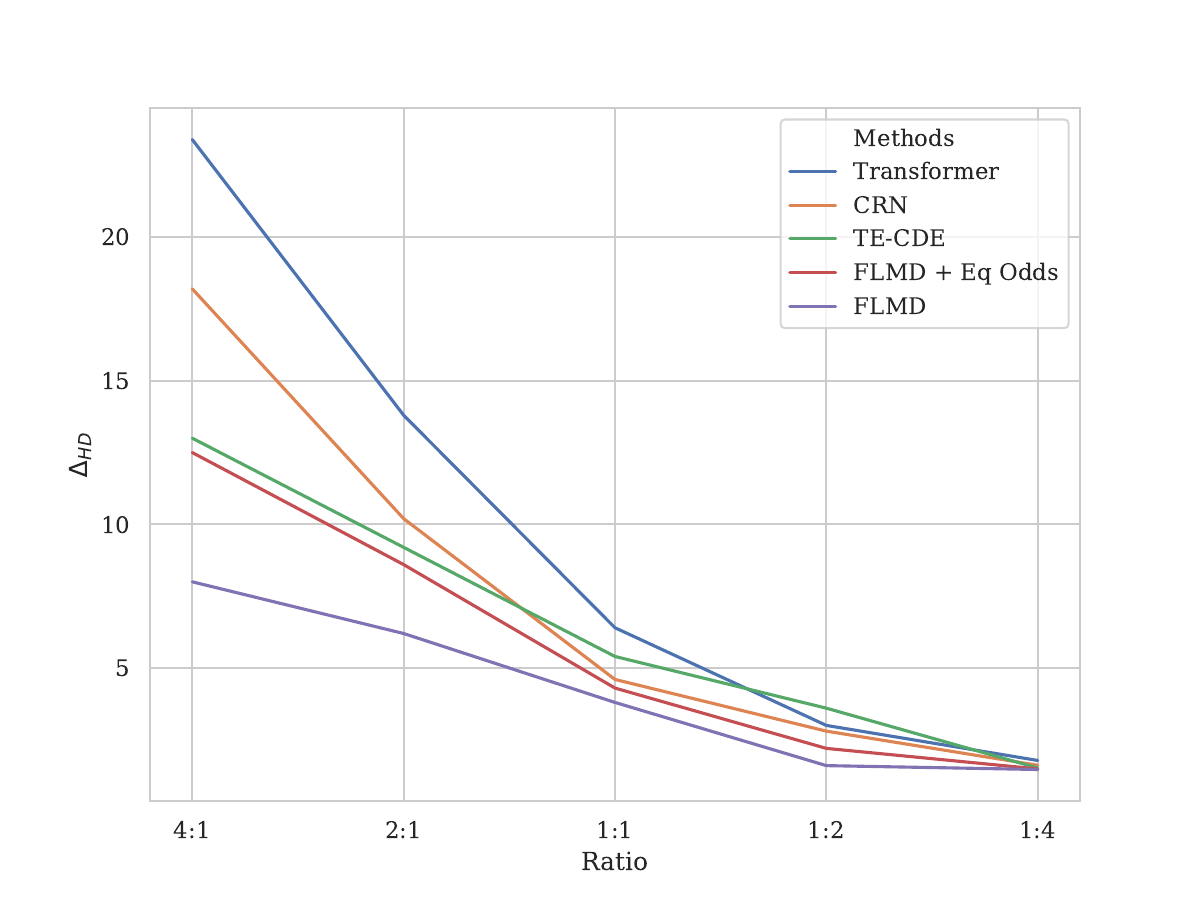}
        \caption{} 
        \label{fig:36}
    \end{subfigure}
    \begin{subfigure}[]{0.23\textwidth}
        \centering
\includegraphics[width=\linewidth,trim=8 13 57 51,,clip]{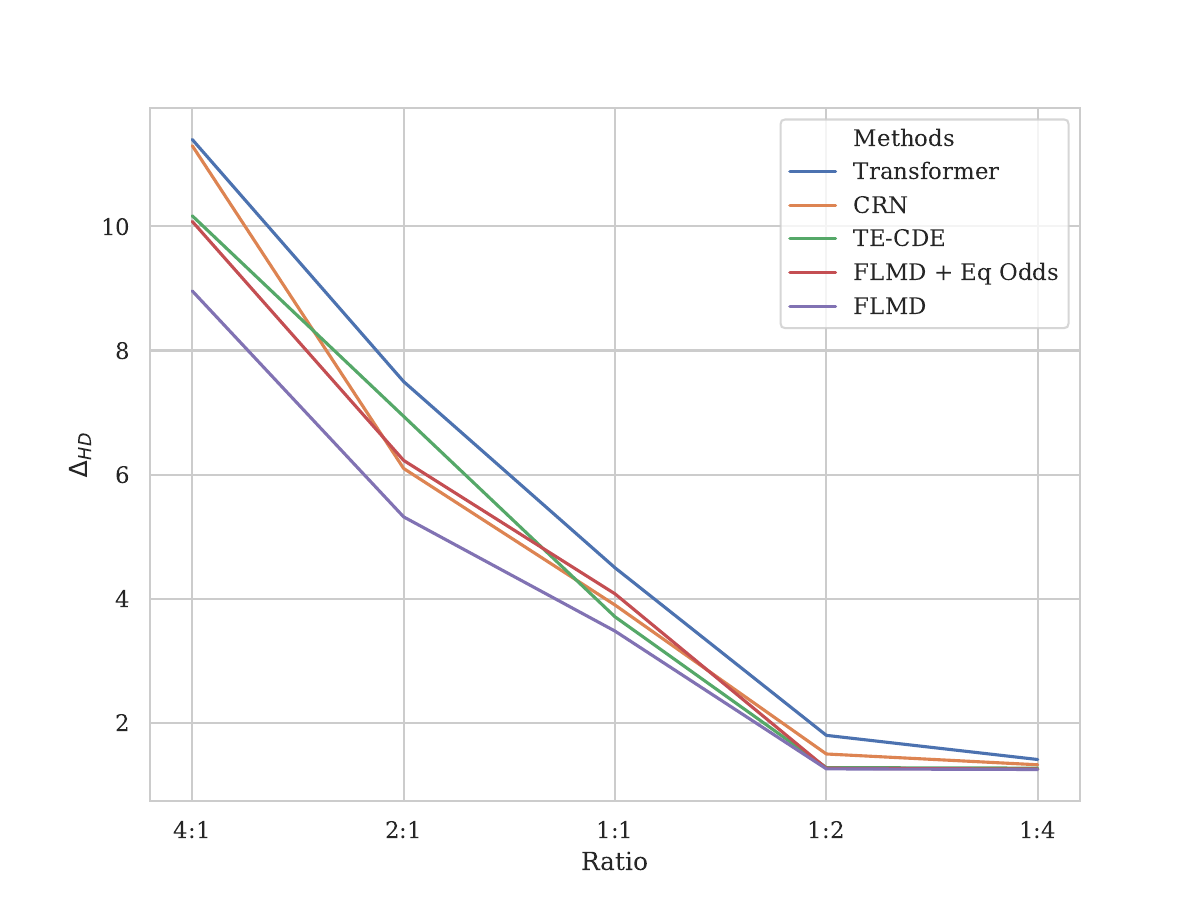}
        \caption{} 
        \label{fig:4237}
    \end{subfigure}
    \begin{subfigure}[]{0.23\textwidth}
        \centering
\includegraphics[width=\linewidth,trim=8 13 57 51,,clip]{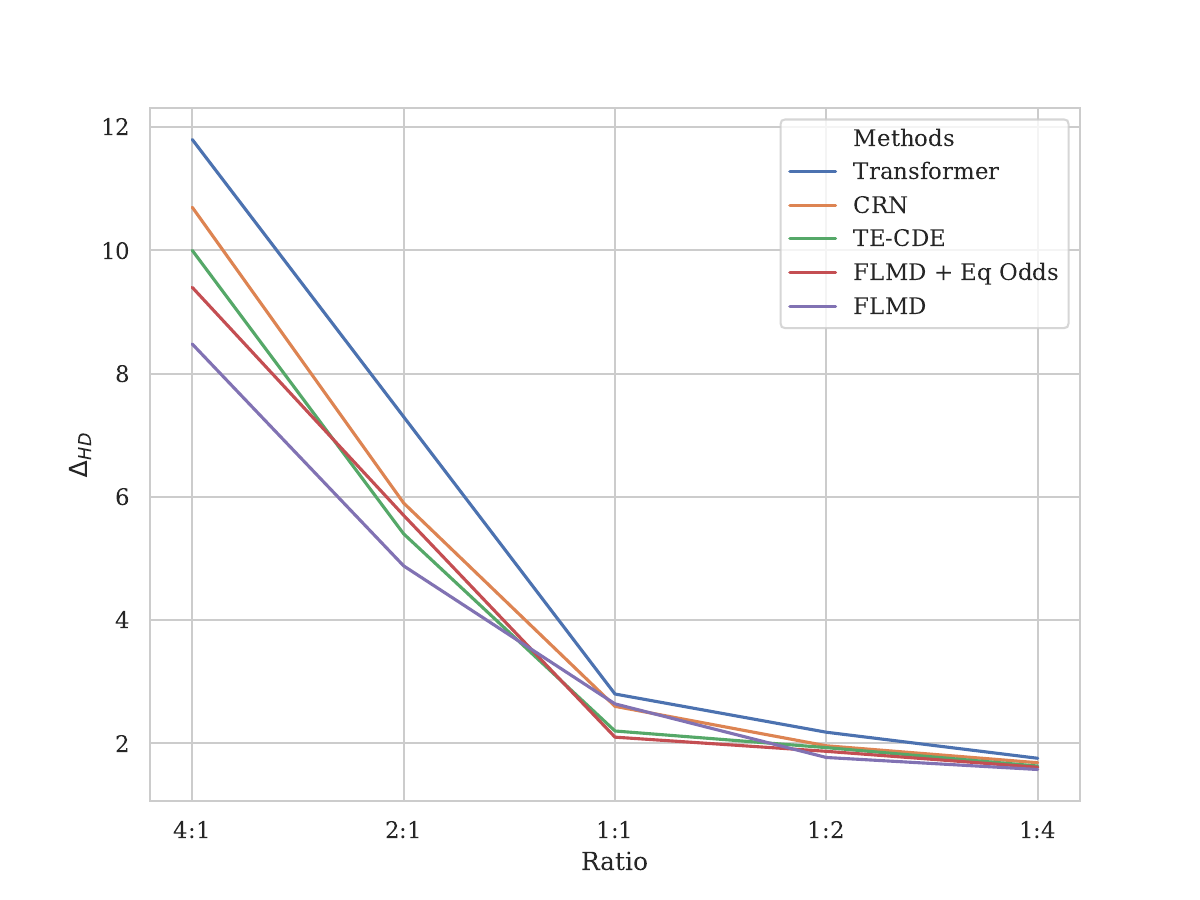}
        \caption{} 
        \label{fig:38}
    \end{subfigure}    
\caption{Accuracy (AUC) and fairness ($\Delta_{HD}$) when the ratio of patient ethnicity (non-White: White) is decided before training. (a)-(d) are comparison of model accuracy and (e)-(h) are comparison of model fairness. (a), (b), (e), (f) are experiments on Hip dataset and (c), (d), (g), (h) are experiments on Knee dataset. (a), (c), (e), (g) are results on PLoS prediction and (b), (d), (f), (h) are results on Readmission prediction.
}  
\label{fig:3}
\end{figure*}

\subsection{Analysis of Generalization Ability (\textbf{Q}2)}

\paragraph{Disturbed data generation and training}
The aim of this experiment is to evaluate the robustness and generalization ability of the model. To achieve this, the data was divided into two datasets: the original dataset and the disturbed dataset. Within the disturbed dataset, certain demographics such as race/ethnicity, gender, and insurance status of each patient were altered.
During the training process, data from both the original and disturbed datasets were used to train the model. The proportions of the original data in the training set were varied as 0.2, 0.4, 0.6, 0.8, and 1.0, corresponding to the horizontal axis of the line plots. Subsequently, the model's performance was evaluated on an independent test set, and both accuracy and fairness metrics were recorded. This approach allowed for the assessment of how well the model performed in different training scenarios and how it maintained fairness in the presence of disturbances in the data.

Figure \ref{tab:q2} shows the performance in terms of accuracy and fairness when the training dataset is disturbed.
In subfigures (a)-(d) of Figure \ref{tab:q2}, where accuracy is measured by the AUC, FLMD demonstrates higher accuracy than the other baselines across different proportions of original training data. On the other hand, in subfigures (e)-(h) of Figure \ref{tab:q2}, where fairness is measured by the health disparity ($\Delta_{HD}$), FLMD exhibits lower disparity compared to the baselines, indicating a fairer model.
Based on these observations, FLMD shows superior performance in terms of both accuracy and fairness when the training distribution differs from the test distribution, outperforming the other baselines across all tasks.
This can be attributed to its higher generalization ability in handling distribution shifts. Although the varying distributions between original and disturbed data pose a challenge, the presence of a shared latent structure implies that a model with high generalization ability can still perform well on disturbed data. These results suggest that the architecture of FLMD aligns well with the latent structure of the model in this scenario.

%% file: Texts/3.3-Q3.tex
\subsection{Analysis on Imbalanced Data (Q2)}  

Figure \ref{fig:3} showcases the performance of all methods in terms of accuracy (AUC) and fairness ($\Delta_{HD}$) as the ratio of training patient ethnicity is manipulated. The figure is generated by controlling the proportion of non-White and White patients in the training data while using a fixed test set to evaluate the performance of all models.

From the observations in the figure, it is discovered that increasing the proportion of White patients in the training set generally leads to improved model performance, given that White patients constitute the majority in the test dataset. However, FLMD consistently maintains the highest accuracy and lowest disparity across different variations in the training data distribution. On the other hand, the performance of the Transformer model, which does not explicitly address fairness, exhibits the most instability and undergoes significant changes as the distribution of training data varies.
Based on these findings, FLMD demonstrates robustness in maintaining high accuracy and fairness even when the distribution of training data is altered. In contrast, the Transformer model's performance proves to be more susceptible to changes in training data distribution due to its lack of explicit fairness considerations.

%% file: Texts/3.4-Q4.tex
\begin{figure}[t]
    \begin{subfigure}[]{0.23\textwidth}
        \centering
\includegraphics[width=\linewidth,trim=16 0 43 39,,clip]{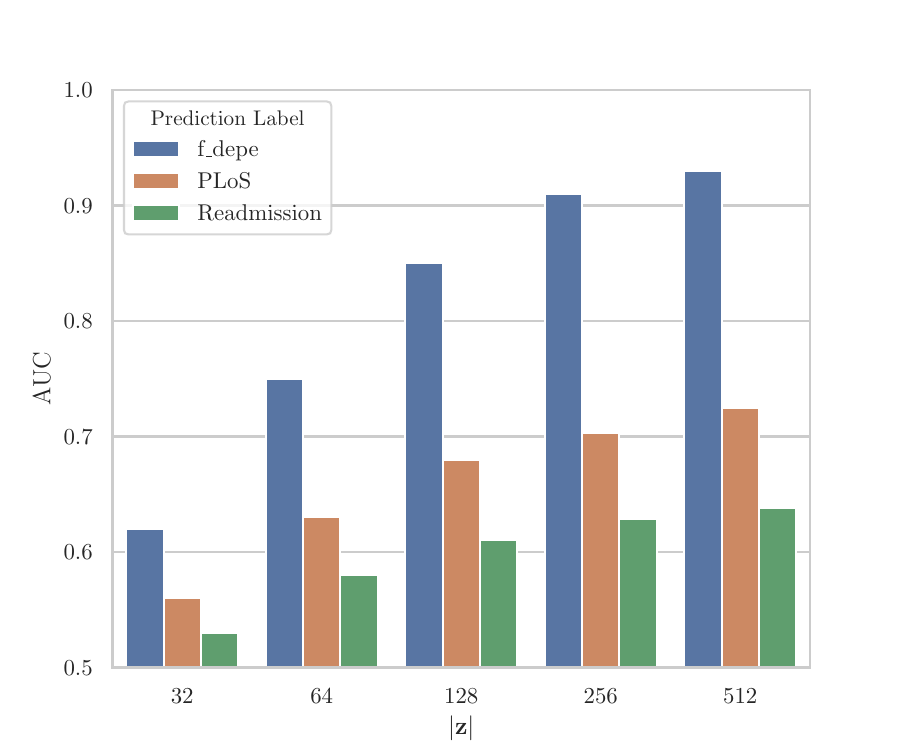}
        \caption{Hip.} 
        \label{fig:41}
    \end{subfigure}
    \begin{subfigure}[]{0.23\textwidth}
        \centering
\includegraphics[width=\linewidth,trim=16 0 43 39,,clip]{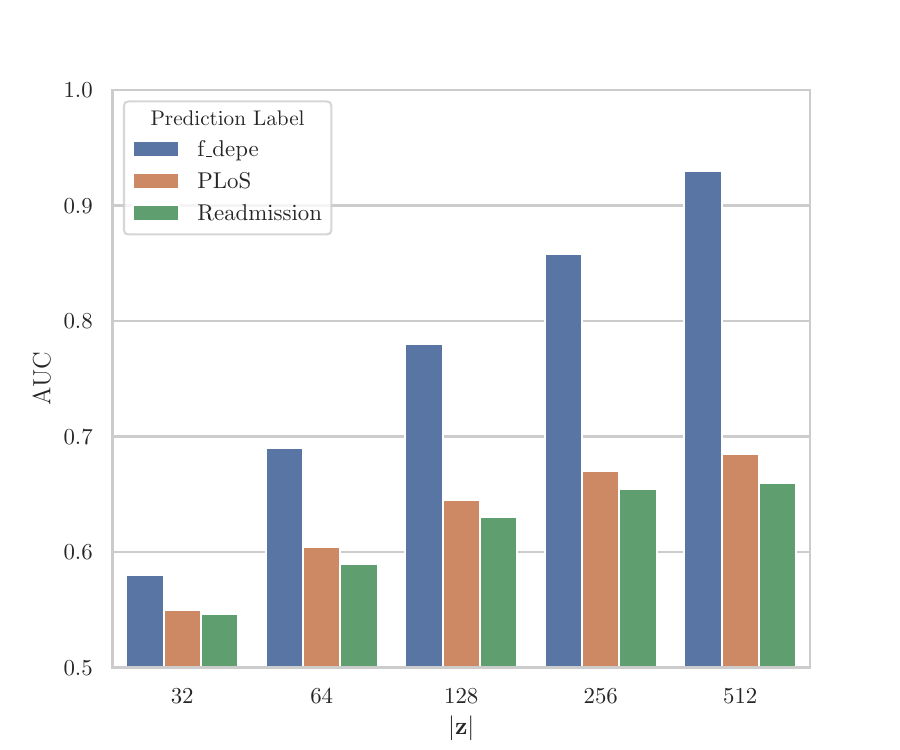}
        \caption{Knee.} 
        \label{fig:42}
    \end{subfigure}
\caption{Prediction Accuracy (AUC) different labels when the length of latent factor $|\bm z|$ varies.
}  
\label{fig:4}
\end{figure} 

\subsection{Analysis of Unobserved Confounder(Q3)} 
We investigate the relationship between unobserved confounders, the latent factor model in FLMD, and modeling performance. To simulate the presence of unobserved confounders that are invisible to humans, a specific observed confounder (financial dependency $f\_depe$) is intentionally made blind to the model. We evaluate the model's ability to predict it to assess whether the model can effectively capture unobserved confounders. 
To enhance the effect of the $f\_depe$ on both features and labels, we manipulate the relationship between them to amplify the causal effects.

\paragraph{Semi-synthetic data generation}
Financial dependency $f\_depe$ is a binary feature that can be categorized into two groups: ``independent" and ``dependent/none". For the features and labels of patients, we use

\begin{align} 
    \bm x_{syn} = 
    \begin{cases}
        \bm x\odot\bm m_1 +\bm b_1 ,& \text{if } f\_depe = 1\\
        \bm x\odot \bm m_2 + \bm b_2    & \text{otherwise}
    \end{cases}
\end{align}

and 
\begin{align} 
    y_{syn} = 
    \begin{cases}
        m_3  y + b_3  ,& \text{if } f\_depe = 1\\
        m_4  y + b_4  & \text{otherwise}
    \end{cases}
\end{align}

to generate synthetic features and labels. Here,  $\odot$ denotes element-wise multiplication and $\bm m_1$, $\bm m_2$, $\bm b_1$, $\bm b_2$, $m_3$, $m_4$, $b_3$, $b_4$ are different synthesis parameters. By choosing appropriate synthesis parameters, the relationship between financial dependency and observational data can be enhanced.

\paragraph{Result Analysis}
Figure \ref{fig:4} presents a comparison of the accuracy (measured by AUC) for predicting financial dependency ($f_depe$), Prolonged Length of Stay (PLoS), and Readmission on synthetic datasets  when varying the length of the latent factor $|\bm z|$. 
For the prediction of $f_depe$, only the first stage of FLMD is deployed and the learned latent factor $|\bm z|$ is used in prediction. 
Subsequently, the second stage of FLMD is deployed to predict PLoS and Readmission.
By assuming that a longer latent factor $\bm z$ signifies a more informative latent factor and an increased model capacity, the experiment investigates the impact of the capacity of the model on the accuracy of capturing unobserved confounders and final predictions. It is observed that as FLMD's first-stage model capacity increases, it demonstrates enhanced capability in capturing the unobserved confounder, resulting in a more useful learned latent confounder $\bm z$ for downstream predictions.




%% file: Texts/4-RelatedWorks.tex
\section{Related Works}
\paragraph{Machine Learning in Healthcare}
Machine learning techniques have gained extensive traction in medical decision-making and healthcare applications, particularly in the processing of EHRs \cite{dai2023diabetic,mou2023automated, li2017blood}. Notably, MedLens~\cite{ye2023medlens} employs data selection and regression methodologies to enhance the predictive accuracy of mortality outcomes. Additionally, FineEHR~\cite{wu2023fineehr} utilizes metric learning and
fine-tuning to refine clinical note embeddings. Moreover, HSGNN~\cite{liu2020heterogeneous} and MHDP~\cite{liu2021medical} employ heterogeneous and hyperbolic graph neural networks to integrate multi-modal data sources effectively.

\paragraph{Fairness in Healthcare}
However, many recent studies have shown that ML models worsen inequity for different patient subgroups \cite{pfohl2021empirical,chen2020ethical,carter2002health}. 
To advance the pursuit of fairness in machine learning models, contemporary research predominantly focuses on two avenues: addressing bias or data incompleteness within the training data \cite{chen2020treating, gaskin2012residential}, or mitigating algorithmic bias within the machine learning model itself \cite{kallus2018residual, jiang2020identifying, mhasawade2021machine}.
While multiple fairness methods and models \cite{pfohl2021empirical,pfohl2022net,liu2022mitigating} can address this inequity, such approaches often require a trade-off between fairness and accuracy. The majority of methods that promote fairness tend to decrease accuracy, thereby reducing the model's overall utility.

\paragraph{Causal Inference in Fairness}
Throughout the evolution of fairness studies, the fairness paradigm/methods rooted in causal inference have increasingly gained prominence and allure.
Counterfactual fairness \cite{kusner2017counterfactual}, which focuses on evaluating the fairness of algorithmic decisions by examining the potential outcomes under different scenarios, offers a more nuanced perspective on fairness considerations.
In this paper, we use another causal inference tool, the deconfounder \cite{wang2019blessings,wang2020towards} to achieve counterfactual fairness further. It uses an unsupervised latent factor model to capture the unobserved confounders before making predictions and further adopts a separate predictive model to derive unbiased results. This model has been widely used in other topics such as recommender systems \cite{wang2020causal} and medical treatment estimation \cite{zhang2019medical}.

%% file: Texts/5-Conclusion.tex
\section{Conclusion}
We propose FLMD, a two-stage framework to learn accurate and generalizable knowledge from longitudinal EHR. FLMD consists of two distinct training stages. First, a deep generative model is used to capture the unobserved confounders (underlying medical factors) in EHR. Second, FLMD combines the latent embeddings with other relevant features as input for a predictive model. By incorporating appropriate fairness criteria,
FLMD ensures that it maintains high prediction accuracy while simultaneously minimizing health disparities.
Experiments are conducted to evaluate the performance of FLMD. We conclude that FLMD can effectively mitigate health disparity and maintain its effectiveness on disturbed or imbalanced data.
We also discover that improving the capacity of the latent factor model can also improve the performance in downstream prediction, which further explains the mechanism of FLMD.

\section{Acknowledgement}
This work is supported in part by NSF under grant  III-2106758.